\documentclass[a4paper,fleqn]{cas-dc}

\usepackage[authoryear]{natbib}
\usepackage{gensymb}
\usepackage{float}
\usepackage{placeins}
\def\tsc#1{\csdef{#1}{\textsc{\lowercase{#1}}\xspace}}
\tsc{WGM}
\tsc{QE}
\tsc{EP}
\tsc{PMS}
\tsc{BEC}
\tsc{DE}

\begin{document}
\let\WriteBookmarks\relax
\def\floatpagepagefraction{1}
\def\textpagefraction{.001}

\shorttitle{GeoSeg-OV: Bridging Geospatial Gaps with Structural Guidance for Open-Vocabulary Remote Sensing Segmentation}

\shortauthors{Ruizhong Liu et~al.}

\title [mode = title]{GeoSeg-OV: Bridging Geospatial Gaps with Structural Guidance for Open-Vocabulary Remote Sensing Segmentation}

\author[1,2]{Ruizhong Liu}[
  style=chinese,
  auid=000,
  bioid=1,
  prefix=,
  orcid=0009-0000-8029-1140]
\ead{rzliu1026@gmail.com}
\credit{Writing - Original draft, Methodology, Software, Data curation, Visualization}

\author[3]{Tingzhang Luo}[
  style=chinese,
  auid=000,
  bioid=2,
  prefix=,
  orcid=0009-0008-4823-5972]
\ead{luotz.gm@gmail.com}
\credit{Methodology, Software, Validation}

\author[4]{Zaiyan Zhang}[
  style=chinese,
  auid=000,
  bioid=1,
  prefix=,
  orcid=0009-0001-7376-7101]
\ead{zzaiyan@whu.edu.cn}
\credit{Methodology, Software, Data curation, Visualization}

\author[5]{Jundong Chen}[
  style=chinese,
  auid=000,
  bioid=1,
  prefix=,
  orcid=0009-0002-3227-5597]
\ead{jundong-chen@biwako.shiga-u.ac.jp}
\credit{Validation, Software, Investigation}

\author[6]{Hongruixuan Chen}[
  style=chinese,
  auid=000,
  bioid=1,
  prefix=,
  orcid=0000-0003-0100-4786]
\ead{qschrx@gmail.com}
\credit{Methodology, Visualization, Writing - Review \& Editing}

\author[1]{Shaoguang Huang}[
  style=chinese,
  auid=000,
  bioid=3,
  prefix=,
  orcid=0000-0001-5439-5018]
\ead{huangshaoguang@cug.edu.cn}
\cormark[1]
\credit{Conceptualization, Supervision, Funding acquisition, Writing - Review \& Editing}

\author[1]{Hongyan Zhang}[
  style=chinese,
  auid=000,
  bioid=6,
  prefix=,
  orcid=0009-0000-5074-0873]
\ead{zhanghongyan@cug.edu.cn}
\credit{Conceptualization, Supervision, Funding acquisition, Writing - Review \& Editing}

\affiliation[1]{organization={School of Computer Science, China University of Geosciences},
  city={Wuhan},
  citysep={}, %
  postcode={430074},
  state={Hubei},
  country={China}}

\affiliation[2]{organization={Systems Hub, The Hong Kong University of Science and Technology (Guangzhou)},
  city={Guangzhou},
  citysep={}, %
  postcode={511453},
  state={Guangdong},
  country={China}}

\affiliation[3]{organization={Department of Computer Science, City University of Hong Kong},
  city={Hong Kong},
  country={China}}

\affiliation[4]{organization={School of Geodesy and Geomatics, Wuhan University},
  city={Wuhan},
  citysep={}, %
  postcode={430079},
  state={Hubei},
  country={China}}

\affiliation[5]{organization={Data Science and AI Innovation Research Promotion Center, Shiga University},
  city={Hikone},
  citysep={}, %
  postcode={522-8522},
  state={Shiga},
  country={Japan}}

\affiliation[6]{organization={RIKEN Center for Advanced Intelligence Project (AIP), RIKEN},
  city={Chuo City},
  citysep={}, %
  postcode={103-0027},
  state={Tokyo},
  country={Japan}}

\cortext[1]{Corresponding author}

\begin{abstract}
Open-vocabulary remote sensing segmentation has recently emerged as a promising paradigm that enables pixel-level recognition of arbitrary categories specified by natural language, including classes unseen during training. However, geospatial domain shifts caused by heterogeneous regions, spatial resolutions, and acquisition platforms weaken visual--text matching and limit cross-dataset generalization. Recent attempts have begun to incorporate auxiliary vision foundation models (VFMs), typically coupling their features with text embeddings as additional matching evidence. However, this strategy may introduce inconsistent matching signals while leaving the structure-sensitive representations of VFMs insufficiently exploited. We therefore propose GeoSeg-OV, which decouples auxiliary VFM features from visual--text matching and repurposes them as structural guidance for cost aggregation and decoding. GeoSeg-OV constructs an orientation-robust cost volume from multi-rotation CLIP features, while a frozen VFM extracts multi-scale structure-sensitive features in parallel. We propose Structure-Guided Aggregation (SGA), which integrates cost tokens and CLIP semantic guidance with VFM-derived pairwise structural biases for coherent spatial propagation, followed by text-conditioned class-wise reasoning. We further introduce Cost-Aware Decoding (CAD) to adaptively refine and fuse multi-scale semantic and structural guidance based on the current decoder context. On the global High-Resolution Land Cover (HRLC) benchmark spanning seven datasets across six continents, GeoSeg-OV outperforms the state-of-the-art by +2.5 and +2.7 average mIoU under two training settings. A large-scale zero-shot case study further demonstrates its generalization across geographic domains and category systems without target-domain annotations or retraining. Code and benchmark are available at \url{https://github.com/zzaiyan/GeoSeg-OV}.
\end{abstract}

\begin{highlights}
\item GeoSeg-OV repurposes auxiliary vision foundation models (VFMs) from text matching to structural guidance.
\item Structure-Guided Aggregation (SGA) integrates semantic--structural spatial aggregation with class-wise reasoning.
\item Cost-Aware Decoding (CAD) enables decoder-conditioned fusion of multi-scale semantic and structural cues.
\item A global HRLC benchmark spans seven datasets across six continents.
\item Large-scale zero-shot mapping requires no target annotations or model retraining.
\end{highlights}


\maketitle

\section{Introduction}

Semantic segmentation assigns a category label to each pixel of an image~\citep{long2015fully,chen2017deeplab,wang2022unetformer} and supports diverse remote sensing applications~\citep{chen2026multimodal,zhang2026ecrformer}, ranging from land use management and urban monitoring to environmental change detection and disaster response. Conventional remote sensing segmentation models, however, are trained on fixed category sets~\citep{liu2024crossmatch,ma2025novel,li2026progressive}, limiting their adaptability to evolving observation demands and previously unseen semantic concepts. Open-vocabulary semantic segmentation (OVSS)~\citep{zhang2023simple,liang2023open} removes this constraint by enabling pixel-level recognition of arbitrary categories described in natural language, exploiting the alignment between visual and textual representations learned by vision--language models~\citep{radford2021learning,zhang2026coreuir}. Although natural-image OVSS has advanced rapidly~\citep{luo2023segclip,yu2023convolutions}, extending it to remote sensing remains an open challenge, especially regarding cross-dataset generalization across heterogeneous land cover imagery, whether through training-free approaches~\citep{li2025annotation,li2025segearth} or trainable frameworks~\citep{cao2025open,ye2025towards,li2026exploring}.

\begin{figure*}[!t]
    \centering
    \includegraphics[width=\textwidth]{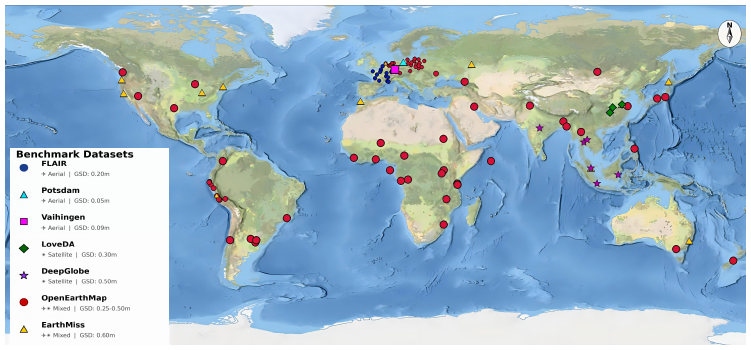}
    \caption{Geographic distribution of the benchmark datasets used in this work. The benchmark spans over 90 cities across 6 continents and covers heterogeneous acquisition platforms, including satellite, aerial, and mixed sources. Ground sampling distances range from 0.05\,m to 0.60\,m, reflecting the multi-source domain gap arising from geographic diversity, platform variation, and resolution differences in high-resolution land cover segmentation.}
    \label{fig:benchmark}
\end{figure*}

Adapting open-vocabulary segmentation to remote sensing introduces challenges beyond the standard image-to-pixel gap commonly encountered in OVSS methods~\citep{jia2021scaling,hu2024reclip}. The major difficulty is the \emph{geospatial gap}: as illustrated in Fig.~\ref{fig:benchmark}, remote sensing imagery is acquired from heterogeneous platforms (satellites, manned aircraft, and drones) whose sensor characteristics, ground sampling distances (0.05\,m to 0.60\,m in our benchmark), and geographic contexts span continents and climate zones. As a result, the same category can appear entirely different across datasets~\citep{li2022breaking,chen2025bright}; for example, a ``building'' captured by a satellite at 0.5\,m resolution over Africa differs in color, texture, roof material, and surrounding context from one imaged by a drone at 0.05\,m over Europe. An effective open-vocabulary method must therefore generalize not only to unseen classes but also to new datasets characterized by shifts in acquisition conditions, spatial resolution, and geographic context, a requirement that is particularly difficult to satisfy for methods that propagate semantic information based on how pixels look rather than how they are spatially organized.

\begin{figure*}[!t]
    \centering
    \includegraphics[width=0.97\textwidth]{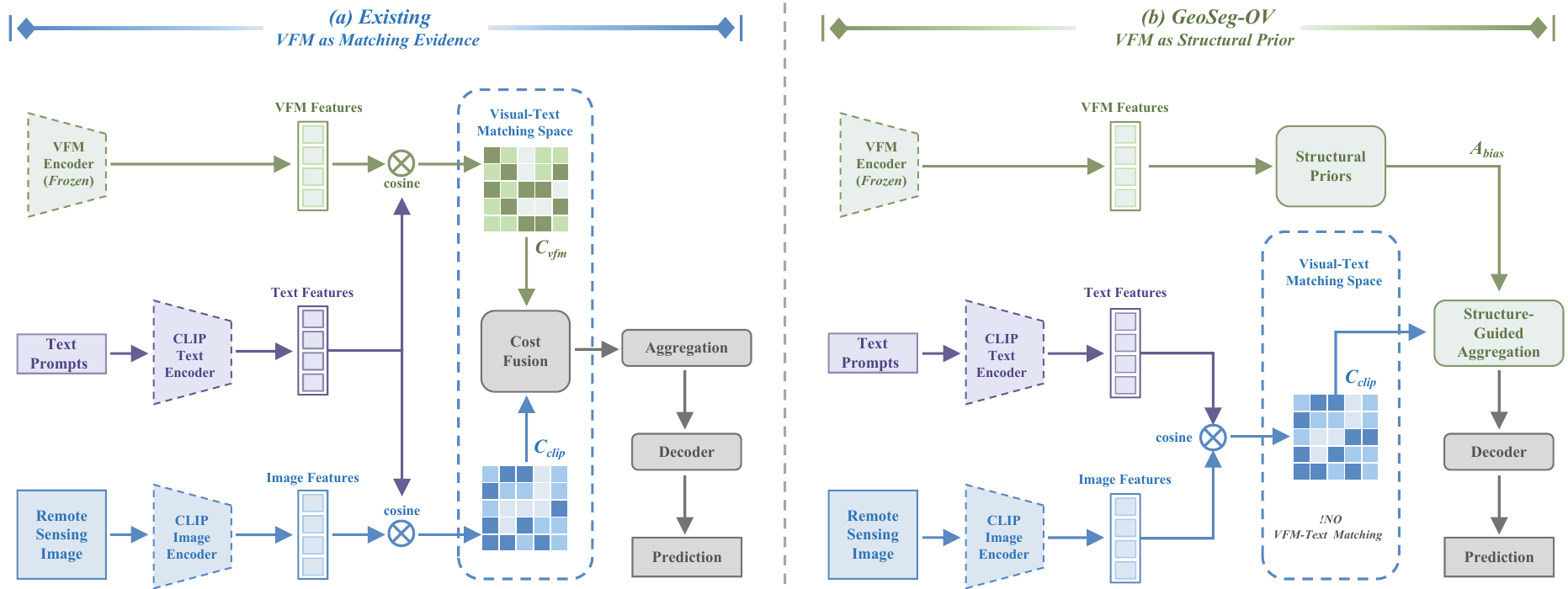}
    \caption{Comparison of two paradigms for utilizing an auxiliary VFM.
    \textbf{(a)}~Existing AVTM: VFM features participate in matching to produce a second cost volume $C_{\text{vfm}}$, which is fused with $C_{\text{clip}}$.
    \textbf{(b)}~Our SGA: the frozen VFM stays outside matching and instead produces a structural bias $A_{\text{bias}}$ that governs cost aggregation of the CLIP-only cost volume.}
    \label{fig:paradigm}
    \vspace{-1em}
\end{figure*}

Among trainable OVSS approaches, the cost aggregation paradigm introduced by CAT-Seg~\citep{cho2024cat} provides an effective framework for open-vocabulary generalization. By aggregating visual--text similarity scores rather than directly decoding image features, it better preserves the CLIP alignment required for recognizing unseen categories~\citep{radford2021learning,luo2026stroke}. However, both cost construction and spatial aggregation in this paradigm are primarily driven by CLIP-derived signals. Under substantial geospatial domain shifts, these signals may become less reliable, producing inconsistent cost patterns for the same category across datasets.
Recently, several remote sensing methods have sought to improve this paradigm by incorporating auxiliary vision foundation models (VFMs), such as domain-adapted DINO encoders~\citep{ye2025towards,li2026exploring}. These methods follow what we refer to as the Auxiliary Visual--Text Matching (AVTM) paradigm: VFM features are correlated with CLIP text embeddings to construct an additional cost volume, treating the auxiliary encoder as a supplementary visual--text matcher, as illustrated in Fig.~\ref{fig:paradigm} (a). However, auxiliary VFMs without explicit visual--text alignment may produce less reliable similarities with CLIP text embeddings, particularly under substantial domain shifts. Consequently, the resulting cost volume may provide limited complementary information or introduce signals that are inconsistent with the original CLIP matching space. In contrast, these VFMs are effective at capturing structure-sensitive cues, including region coherence, boundary organization, and spatial layout, which may be more transferable across variations in appearance. This motivates us to use VFM features outside the visual--text matching process as an independent structural reference for guiding the cost aggregation of CLIP matching evidence. We refer to this paradigm as Structure-Guided Aggregation (SGA), as illustrated in Fig.~\ref{fig:paradigm} (b).

\begin{figure}[t]
    \centering
    \includegraphics[width=\linewidth]{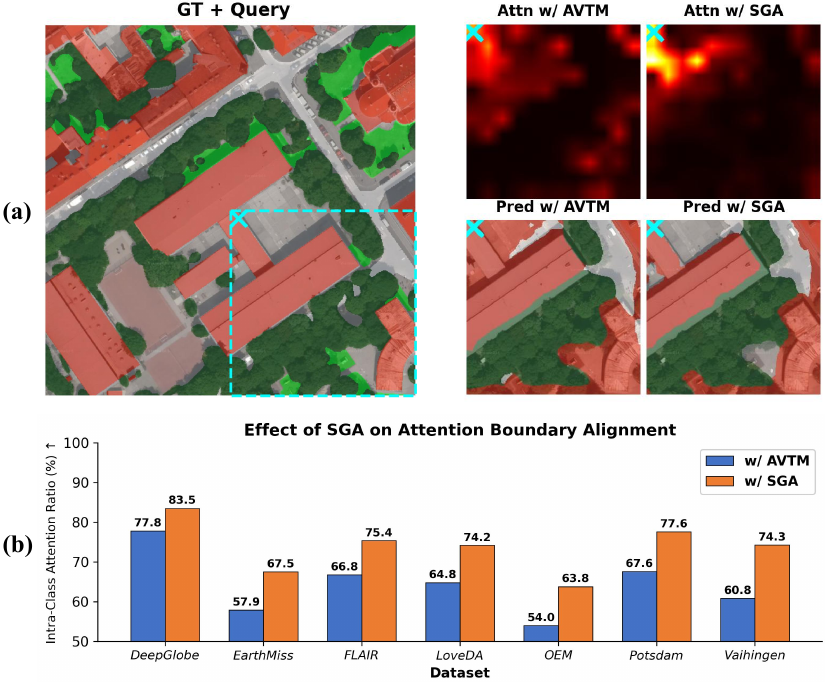}
    \caption{
    Attention comparison between AVTM and SGA.
    (a) For a query on a building (cyan cross), AVTM attention spreads beyond the 
    object boundary and mispredicts, while SGA stays within the region.
    (b) Intra-class attention ratio (attention assigned to the query's own class, higher 
    is better) on seven datasets.
    }
    \label{fig:sga_case}
\end{figure}

We empirically compare the AVTM and SGA paradigms in Fig.~\ref{fig:sga_case}. As shown in Fig.~\ref{fig:sga_case} (a), under AVTM, cost-token attention may extend beyond object boundaries, causing matching evidence to be exchanged between structurally unrelated regions. This suggests that introducing an additional cost volume alone does not directly address unreliable spatial propagation. In contrast, SGA transforms auxiliary VFM features into a pairwise structure-sensitive bias and injects it into the attention computation, providing complementary guidance on which positions should exchange matching evidence. Consequently, the resulting attention is more concentrated within coherent regions and better aligned with object boundaries, leading to a more accurate segmentation result in the illustrated example. Fig.~\ref{fig:sga_case} (b) further quantifies this comparison across seven datasets. SGA consistently achieves a higher intra-class attention ratio than AVTM, supporting the effectiveness of using VFM features to guide cost aggregation rather than treating them as additional visual--text matching signals.

Based on this observation, we propose GeoSeg-OV, a unified structure-guided framework for open-vocabulary remote sensing segmentation. Given an input image and category descriptions, GeoSeg-OV employs multi-orientation CLIP encoding to construct an orientation-robust visual--text cost volume, while a frozen auxiliary VFM extracts multi-scale structure-sensitive features in parallel. We propose Structure-Guided Aggregation (SGA) to progressively refine the cost representation across spatial and category dimensions. SGA integrates cost tokens and CLIP semantic guidance with VFM-derived pairwise structural biases for coherent spatial propagation, followed by text-conditioned class-wise reasoning. We further introduce Cost-Aware Decoding (CAD) to adaptively refine and fuse multi-scale semantic and structural guidance based on the current decoder context, progressively recovering pixel-level predictions. Since the auxiliary VFM serves only as a structural provider outside the matching space, GeoSeg-OV is robust to different VFM choices and consistently benefits from diverse pretrained representations. In this way, GeoSeg-OV preserves CLIP as the sole source of visual--text matching while fully exploiting complementary VFM features throughout cost aggregation and decoding.

Our main contributions are summarized as follows:
\begin{itemize}
    \item We propose GeoSeg-OV, a unified structure-guided framework that decouples auxiliary VFM features from visual--text matching. GeoSeg-OV preserves CLIP-based cost-volume construction while repurposing frozen VFM features as complementary structural guidance for both aggregation and decoding.

    \item We develop SGA, which integrates semantic--structural spatial propagation with text-conditioned class-wise reasoning to produce coherent and discriminative cost representations. We further propose CAD to adaptively refine and fuse multi-scale semantic and structural guidance according to the current decoder context.

    \item We establish a unified High-Resolution Land Cover (HRLC) benchmark spanning seven datasets across six continents, on which GeoSeg-OV achieves state-of-the-art cross-dataset performance. A large-scale zero-shot case study further demonstrates its transferability across geographic domains and category systems without target-domain annotations or retraining.
\end{itemize}

\section{Related Work}

\subsection{Open-Vocabulary Semantic Segmentation}

OVSS aims to assign pixel-level labels for arbitrary categories specified through text descriptions~\citep{li2026exploringwater,li2026maris}, moving beyond the closed-set assumption of traditional segmentation~\citep{radford2021learning}. Vision--language models, particularly CLIP~\citep{radford2021learning}, make this possible by aligning visual and textual representations in a shared embedding space. Existing OVSS methods can be broadly categorized into two-stage and single-stage approaches.

Two-stage methods first generate class-agnostic mask proposals and then classify each proposal against text embeddings. OpenSeg~\citep{ghiasi2022scaling} learns region-level visual embeddings by correlating local image regions with text descriptions. OVSeg~\citep{liang2023open} fine-tunes CLIP on region-text pairs to improve mask-text alignment. ODISE~\citep{xu2023open} uses pre-trained Stable Diffusion to produce high-quality class-agnostic masks, which are subsequently classified by CLIP. While effective, the two-stage pipelines depend on external proposal generation, introducing additional complexity and potential domain sensitivity.

Single-stage methods directly predict segmentation masks conditioned on text prompts. SED~\citep{xie2024sed} proposes a simple encoder-decoder architecture that bridges CLIP features to dense prediction. SAN~\citep{xu2023side} introduces side adapter networks to extract multi-scale features from a frozen CLIP encoder, preserving open-vocabulary alignment while enabling dense output. CAT-Seg~\citep{cho2024cat} introduces a distinct cost aggregation paradigm: it computes dense cosine similarities between CLIP image and text embeddings to construct a vision--language cost volume, which is then aggregated through Swin-based transformer layers. By operating on matching costs rather than task-specific decoded features, this framework preserves the capacity of CLIP to recognize unseen categories without degrading the pre-trained alignment. FC-CLIP~\citep{yu2023convolutions} further demonstrates that keeping the CLIP image encoder entirely frozen benefits unseen-class recognition, reinforcing the importance of alignment preservation. The cost aggregation paradigm has since become the dominant framework for open-vocabulary segmentation in remote sensing~\citep{cao2025open,ye2025towards,li2026exploring}.

\subsection{Open-Vocabulary Remote Sensing Segmentation}

Extending OVSS to remote sensing is challenging due to the geospatial gap between natural and overhead imagery~\citep{zhang2025multi,zhang2026task,luo2026cross}. Heterogeneous acquisition platforms, sensor modalities, spatial resolutions, and geographic variation introduce distribution shifts that substantially weaken vision--language matching learned from natural images. Recent remote sensing OVSS methods have advanced along both training-free~\citep{li2025annotation,li2025segearth} and trainable~\citep{zermatten2025learning,huang2026reducing} directions. Since trainable methods can learn task-specific cost aggregation beyond what frozen VLM features provide, we focus on this line of work.

OVRS~\citep{cao2025open} is among the first to adapt cost aggregation for open-vocabulary remote sensing segmentation, achieving notable improvements over the natural-image baseline. GSNet~\citep{ye2025towards} introduces a dual-stream architecture that pairs the CLIP encoder with a DINO-based remote sensing backbone, fusing domain-specific representations into cost-map construction through query-guided correlation. RSKT-Seg~\citep{li2026exploring} integrates multi-direction cost aggregation with knowledge transfer from multiple domain-adapted encoders through an efficient fusion transformer, achieving strong performance with faster inference. These methods commonly couple auxiliary features with visual--text matching or cost-map construction to enhance semantic evidence. However, this strategy may introduce inconsistent matching signals under domain shifts while underutilizing the structure-sensitive representations of auxiliary VFMs. We therefore decouple auxiliary VFM features from visual--text matching and reposition them as independent structural priors for cost aggregation and decoding.


\begin{figure*}[t]
    \centering
    \includegraphics[width=\textwidth]{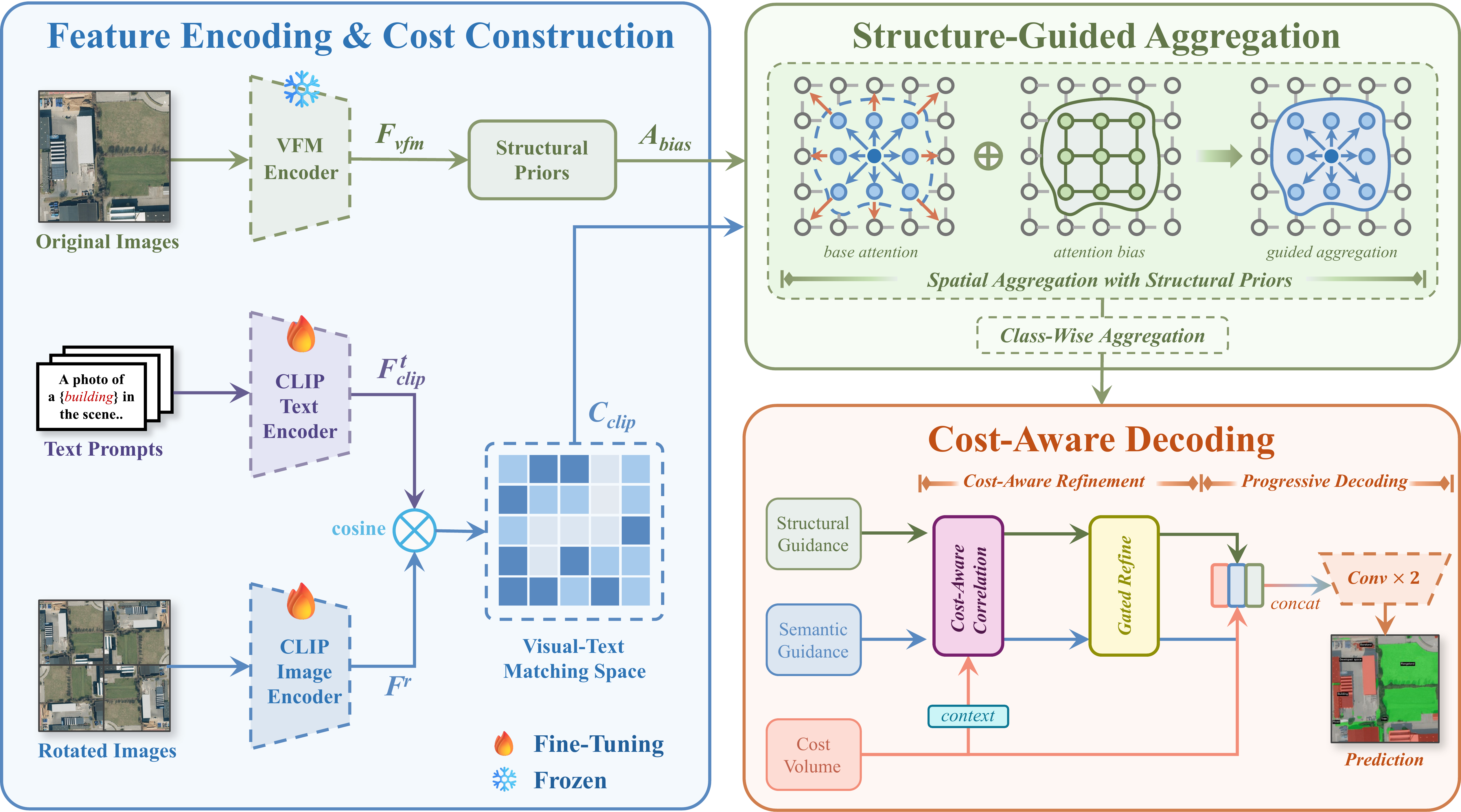}
    \caption{Overall framework of GeoSeg-OV, organized into three stages. 1) \emph{Feature encoding and cost construction}: only CLIP features enter the visual--text matching space to build the cost volume $C_{\mathrm{clip}}$, where the image is encoded under four rotations for orientation robustness, while a frozen auxiliary VFM produces structural features $F_{\mathrm{vfm}}$ that stay outside this matching space. 2) \emph{Structure-Guided Aggregation}: cost tokens and CLIP semantic guidance establish the base spatial affinity, which is complemented by VFM-derived pairwise structural biases to guide evidence propagation within coherent regions. The resulting structure-consistent cost representations are further refined through text-conditioned class-wise reasoning. 3) \emph{Cost-Aware Decoding}: the aggregated cost state conditions the refinement of both semantic and structural guidance, which are then concatenated with the upsampled cost and decoded progressively to full resolution.}
    \label{fig:method}
    \vspace{-1em}
\end{figure*}

\section{Methodology}

Given an input image $I\in\mathbb{R}^{H\times W\times 3}$ and a set of $K$ category descriptions $\mathcal{T}=\{t_k\}_{k=1}^{K}$ provided as free-form text, open-vocabulary remote sensing segmentation assigns a class label to each pixel in $I$ based on $\mathcal{T}$, where the category set may differ between training and inference. In this section, we present GeoSeg-OV, a unified structure-guided framework that decouples auxiliary VFM features from visual--text matching and repurposes them to regulate cost aggregation and enable context-adaptive decoding, as illustrated in Fig.~\ref{fig:method}.


The overall pipeline proceeds as follows. The input image is encoded by the CLIP image encoder under four rotations, and the resulting features are counter-rotated to obtain spatially aligned, orientation-robust representations. These visual features are matched with the CLIP text embeddings to construct a multi-rotation cost volume, which is subsequently projected into cost embeddings. In parallel, a frozen auxiliary VFM extracts multi-scale structure-sensitive features from the original image. The cost embeddings are then progressively refined through alternating spatial and class-wise aggregation. Finally, CAD progressively upsamples the aggregated cost representation, refining both CLIP and VFM guidance features conditioned on the cost state at each scale before fusion. The trainable components are optimized using a per-pixel binary cross-entropy loss.

\subsection{Multi-Rotation Cost Volume Construction}
\label{sec:cost_volume}
GeoSeg-OV constructs the cost volume exclusively from CLIP, retaining it as the sole source of visual--text matching for open-vocabulary recognition. To improve robustness to the diverse object orientations in overhead imagery, we adopt multi-rotation encoding to obtain spatially aligned visual features from multiple views.

Specifically, we encode the input image $I\in\mathbb{R}^{H\times W\times 3}$ under four rotations~\citet{cao2025open}. Let $\Phi_v$ denote the CLIP image encoder and $\mathcal{R}_r(\cdot)$ denote spatial rotation by $r\times90^\circ$. Each rotated image is encoded and then counter-rotated back to the canonical orientation:
\begin{equation}
    F^r = \mathcal{R}_{-r}\!\bigl(\Phi_v(\mathcal{R}_r(I))\bigr),\quad r\in\{0,1,2,3\},
\end{equation}
yielding four spatially aligned feature maps $F^r\in\mathbb{R}^{H_0\times W_0\times D_v}$, where $H_0\times W_0$ is the spatial resolution of the CLIP patch embeddings and $D_v$ is the feature dimension. In parallel, the CLIP text encoder $\Phi_t$ encodes each category description under $P$ prompt templates, producing text embeddings $F_{\mathrm{clip}}^t\in\mathbb{R}^{K\times P\times D_v}$.

The cost volume is constructed as the dense cosine similarity between each rotated visual feature and the text embeddings:
\begin{equation}
    C(r,p,k,u) = \frac{\langle\, F^r(u),\; F_{\mathrm{clip}}^{t}(k,p)\,\rangle}{\|F^r(u)\|_2\;\|F_{\mathrm{clip}}^{t}(k,p)\|_2}\,,\label{eq:cost_volume}
\end{equation}
where $u$ indexes spatial positions, $k$ indexes categories, and $p$ indexes prompt templates. Concatenating across the four rotations and $P$ templates produces the multi-rotation cost volume $C_{\mathrm{clip}}\in\mathbb{R}^{4P\times K\times H_0\times W_0}$. This volume captures orientation-robust matching evidence between every spatial location and every category, serving as input to subsequent structure-guided aggregation.

\subsection{Structure-Guided Aggregation}
\label{sec:sga}

With the cost volume constructed, the next stage refines it into a discriminative representation for dense semantic prediction. The cost volume $C_{\mathrm{clip}}$ encodes per-position, per-category matching scores, but these raw scores are computed independently at each location and might be noisy due to domain shift. The aggregation stage addresses this by refining the cost volume across spatial and class domains.

The cost volume is first projected into a latent representation $X_0\in\mathbb{R}^{B\times C\times K\times H_0\times W_0}$ through a convolutional embedding layer that maps the $4P$ raw similarity channels of each category slice to a $C$-dimensional feature space, where $B$ is the batch size and $C$ is the cost embedding dimension. This lifts the raw similarities into a higher-dimensional space that better supports subsequent transformer processing. The representation is then refined through $L$ layers, each first applying spatial aggregation $\Phi_{\ell}$ to propagate evidence across spatial neighbors, then class-wise aggregation $\Gamma_{\ell}$ to reason about inter-category dependencies at each position, with different guidance signals conditioning each stage:
\begin{equation}
    X_{\ell} = \Gamma_{\ell}\!\bigl(\Phi_{\ell}(X_{\ell-1};\,S_{1},G_{1}),\;E_{t}\bigr), \quad \ell=1,\dots,L,
\end{equation}
where $S_1$ and $G_1$ denote semantic and structural guidance,
respectively (defined in Sec.~\ref{sec:spatial_agg}). For class-wise
conditioning, the $P$ prompt embeddings of each category are first
averaged and $\ell_2$-normalized, and the resulting category embedding
is then projected to the cost-embedding dimension:
\begin{equation}
\begin{aligned}
    \mu_t(k)
    &=
    \frac{1}{P}
    \sum_{p=1}^{P}
    F_{\mathrm{clip}}^{t}(k,p),\\
    \bar{F}_t(k)
    &=
    \frac{\mu_t(k)}
         {\left\|\mu_t(k)\right\|_2},\\
    E_t
    &=
    \mathrm{Proj}_{t}(\bar{F}_t)
    \in \mathbb{R}^{K\times C}.
\end{aligned}
\end{equation}

The key design of SGA lies in how structural priors from the auxiliary VFM are introduced to govern the aggregation process. In the spatial stage, they provide pairwise structure-sensitive biases that regulate evidence propagation across positions; in the class-wise stage, the spatially coherent cost representations produced under structural guidance in turn enable more reliable inter-category reasoning. We detail both stages below.

\subsubsection{Spatial Aggregation with Structural Priors}
\label{sec:spatial_agg}


The raw cost volume contains independently computed matching responses that may be noisy and spatially fragmented. SGA refines these responses by jointly modeling semantic relevance and structural coherence. Specifically, cost tokens and intermediate CLIP features establish the base content affinity, preserving category-aware semantic relationships during spatial interaction. In parallel, a frozen VFM provides pairwise structure-sensitive biases outside the visual--text matching space, constraining evidence propagation according to region organization and object boundaries. By integrating these complementary signals, SGA produces a structure-consistent and semantically discriminative cost representation, with coherent responses within regions and clear separation across boundaries.

Specifically, we first extract multi-scale semantic guidance from intermediate features of the canonical CLIP branch ($r{=}0$):
\begin{equation}
    S_l
    =
    \mathrm{Proj}_s^l
    \!\left(F_{\mathrm{clip}}^l\right),
    \qquad l\in\{1,2,3\},
\end{equation}
where $F_{\mathrm{clip}}^l$ denotes the intermediate CLIP feature at scale $l$, and $\mathrm{Proj}_s^l$ is the corresponding learned convolutional projection. Here, $S_1$ is used for spatial aggregation, while $\{S_2,S_3\}$ are used during progressive decoding.

At the aggregation scale, spatial aggregation is performed within local windows, each containing $N$ cost tokens $x\in\mathbb{R}^{N\times C}$. Let $S_1\in\mathbb{R}^{B\times D_s\times H_0\times W_0}$ denote the projected semantic guidance and $s\in\mathbb{R}^{N\times D_s}$ its windowed representation, where $D_s$ is the semantic guidance dimension. The base content attention is computed as
\begin{equation}
    q=[x;s]W_q,\qquad
    k=[x;s]W_k,\qquad
    v=xW_v,
\end{equation}
\begin{align}
    \mathrm{Attn}_{\mathrm{base}}(i,j)
    &=
    \mathrm{softmax}
    \left(
    \frac{q_i k_j^\top}{\sqrt{d_k}}
    \right),\\
    x_i'
    &=
    \sum_{j=1}^{N}
    \mathrm{Attn}_{\mathrm{base}}(i,j)v_j,
\end{align}
where $W_q,W_k\in\mathbb{R}^{(C+D_s)\times N_h d_k}$, $W_v\in\mathbb{R}^{C\times N_h d_k}$, and $d_k=C/N_h$ is the dimension of each of the $N_h$ attention heads. The projection outputs are reshaped into $N_h$ heads before attention computation, and the head index is omitted for clarity. The cost tokens and CLIP semantic guidance jointly determine the attention affinity, whereas the values are computed solely from the cost tokens to preserve the matching evidence being propagated.

To complement this CLIP-derived affinity, we employ a frozen VFM $\Phi_{\mathrm{vfm}}$ to extract multi-scale structure-sensitive features from the input image:
\begin{equation}
    G_l
    =
    \mathrm{Proj}_g^l
    \!\left(F_{\mathrm{vfm}}^l\right),
    \qquad l\in\{1,2,3\},
\end{equation}
where $F_{\mathrm{vfm}}^l$ denotes the selected intermediate VFM feature at scale $l$, and $\mathrm{Proj}_g^l$ is a learned projection. Here, $G_1\in\mathbb{R}^{B\times D_g\times H_0\times W_0}$ is used for spatial aggregation, while $\{G_2,G_3\}$ are passed to the decoder~(\S\ref{sec:cad}). Importantly, these features remain outside the visual--text matching process and are used only to guide the propagation of the CLIP-based matching evidence.

Let $g\in\mathbb{R}^{N\times D_g}$ denote the windowed structural guidance from $G_1$, where $D_g$ is the structural guidance dimension. We transform $g$ into a head-specific pairwise structure-sensitive bias:
\begin{equation}
    q^s=gW_q^s,\qquad
    k^s=gW_k^s,\qquad
    A_{\mathrm{bias}}(i,j)
    =
    \frac{q_i^s{k_j^s}^{\top}}{\sqrt{d_h}},
\end{equation}
where $W_q^s,W_k^s\in\mathbb{R}^{D_g\times N_h d_h}$, and $d_h$ is the structural dimension of each attention head. The projected features are reshaped into $N_h$ heads to compute head-specific biases. Unlike the base affinity jointly determined by the cost tokens and CLIP semantic guidance, $A_{\mathrm{bias}}$ provides an independent pairwise term derived solely from the auxiliary VFM features.

We then incorporate the structural bias into the attention logits before softmax:
\begin{align}
    \mathrm{Attn}(i,j)
    &=
    \mathrm{softmax}
    \left(
    \frac{q_i k_j^\top}{\sqrt{d_k}}
    +
    A_{\mathrm{bias}}(i,j)
    \right),\\
    x_i'
    &=
    \sum_{j=1}^{N}
    \mathrm{Attn}(i,j)v_j.
\end{align}
In this way, the CLIP-derived affinity provides semantic guidance for spatial interaction, while the VFM-derived bias introduces complementary structure-sensitive information to regulate where matching evidence is propagated. In practice, each aggregation layer contains two consecutive Swin Transformer blocks with regular and shifted windows~\citep{liu2021swin}, and $A_{\mathrm{bias}}$ is injected into both blocks. Since the values are computed exclusively from the cost tokens, SGA guides spatial information flow without directly incorporating VFM features into the propagated representation, thereby improving regional coherence and reducing information exchange across object boundaries.

\subsubsection{Class-Wise Aggregation with Text Conditioning}
\label{sec:class_agg}

The preceding spatial stage, guided by structural priors, produces a structure-consistent cost representation for each category independently. However, it does not explicitly model relationships among categories. Consequently, a spatial position may exhibit competing responses to semantically related categories, such as ``tree'' and ``low vegetation'' or ``building'' and ``impervious surface''. To reduce this ambiguity, class-wise aggregation enables category-specific cost embeddings at the same spatial position to interact and jointly refine their responses.

Specifically, for each spatial position $i$, the cost embeddings of all $K$ categories are gathered and processed by a linear attention layer~\citep{katharopoulos2020transformers} conditioned on the text embeddings $E_t$:
\begin{equation}
    X''(i,:)
    =
    \mathrm{LinearAttn}
    \!\left(
    [X'(i,:);E_t]
    \right),
\end{equation}
where $X'(i,:)\in\mathbb{R}^{K\times C}$ denotes the spatially aggregated cost embeddings at position $i$. The text embeddings are incorporated into the queries and keys to condition category interactions on their semantic relationships. Unlike spatial aggregation, attention here operates over the category dimension, allowing each category to interact with all other categories at the same spatial position. We adopt linear attention to maintain linear complexity with respect to $K$, facilitating inference with varying vocabulary sizes.

Alternating spatial and class-wise aggregation over $L$ layers (with $L{=}2$ by default) allows the two forms of reasoning to complement each other: spatial aggregation improves regional coherence within each category, while class-wise aggregation reduces ambiguity among competing categories. The resulting representation $X_L$ is subsequently passed to the decoder.

\subsection{Cost-Aware Decoding}
\label{sec:cad}

After cost aggregation at low resolution, the decoder progressively recovers pixel-level predictions with the aid of multi-scale guidance features $\{S_l,G_l\}$. The semantic guidance $S_l$ provides complementary semantic information, while the structural guidance $G_l$ contributes cues related to region coherence and object boundaries. A straightforward strategy is to uniformly concatenate these features with the upsampled decoder representation. However, this strategy does not account for the current decoding context, although the relevance of different guidance cues may vary across spatial locations and decoding stages.

To address this limitation, we propose CAD, which uses a category-shared context derived from the current decoder representation to refine the semantic and structural guidance before fusion. Specifically, this context interacts with each guidance stream to generate a spatial gate for decoder-conditioned feature refinement. The resulting guidance features are then fused with the upsampled decoder representation for progressive prediction.

\subsubsection{Cost-Aware Refinement}
\label{sec:cad_module}

The final aggregated representation
$X_L\in\mathbb{R}^{B\times C\times K\times H_0\times W_0}$
is first reshaped by folding the category dimension into the batch dimension:
\begin{equation}
    D_1
    =
    \mathrm{reshape}(X_L)
    \in
    \mathbb{R}^{(BK)\times C\times H_0\times W_0}.
\end{equation}
Refinement is then performed at each decoding scale $l\in\{2,3\}$. Specifically, the decoder state from the preceding stage is upsampled using a transposed convolution:
\begin{equation}
    \widetilde{D}_l
    =
    \mathrm{ConvT}_{\uparrow 2}(D_{l-1})
    \in
    \mathbb{R}^{(BK)\times C_{u,l}\times H_l\times W_l},
\end{equation}
where $C_{u,l}$ denotes the channel dimension after upsampling, and $H_l\times W_l$ is the target resolution at stage $l$.

To derive a category-shared decoder context, we first recover the category dimension and then perform mean pooling over all categories:
\begin{equation}
\begin{aligned}
    \widetilde{D}_l^{\mathrm{cat}}
    &=
    \mathrm{reshape}_{B,K}
    \left(
    \widetilde{D}_l
    \right)
    \in
    \mathbb{R}^{B\times K\times C_{u,l}\times H_l\times W_l},\\
    \overline{D}_l
    &=
    \frac{1}{K}
    \sum_{k=1}^{K}
    \widetilde{D}_l^{\mathrm{cat}}(:,k,:,:,:)
    \in
    \mathbb{R}^{B\times C_{u,l}\times H_l\times W_l}.
\end{aligned}
\end{equation}
Mean pooling provides a permutation-invariant and size-normalized summary of the current category set. We use $\overline{D}_l$ as a category-shared decoder context rather than interpreting it as an explicit confidence or uncertainty estimate.

For each guidance stream $U_l\in\{S_l,G_l\}$, we project $\overline{D}_l$ to the corresponding guidance dimension and compute a relevance gate through its interaction with $U_l$:
\begin{equation}
\begin{aligned}
    P_l
    &=
    \mathrm{GN}
    \left(
    \mathrm{Conv}_{1\times1}
    \left(
    \overline{D}_l
    \right)
    \right),\\
    R_l
    &=
    \sigma
    \left(
    P_l\odot U_l
    \right),
\end{aligned}
\end{equation}
where $\sigma$ denotes the sigmoid function and $\odot$ denotes element-wise multiplication. For clarity, the guidance-stream index on $P_l$, $R_l$, and the corresponding operators is omitted; separate parameters are used for the semantic and structural streams. The resulting gate is jointly conditioned on the decoder context and the corresponding guidance feature.

The gated guidance is then locally refined and added to the original feature through a residual connection:
\begin{equation}
    \widehat{U}_l
    =
    U_l
    +
    \mathrm{GN}
    \left(
    \mathrm{PW}
    \left(
    \mathrm{DW}_{3\times3}
    \left(
    R_l\odot U_l
    \right)
    \right)
    \right),
\end{equation}
where $\mathrm{DW}_{3\times3}$ and $\mathrm{PW}$ denote depthwise and pointwise convolutions, respectively. The residual connection preserves the original guidance, while the gated branch introduces decoder-conditioned local refinement.

\subsubsection{Progressive Decoding}
\label{sec:decoder}

Cost-aware refinement is applied independently to $S_l$ and $G_l$, producing the decoder-conditioned guidance features $\widehat{S}_l$ and $\widehat{G}_l$, respectively. The two streams are kept separate because they provide complementary information: semantic guidance supports category discrimination, while structural guidance contributes to regional coherence and boundary delineation.

The refined guidance features are broadcast across the $K$ categories, folded into the batch dimension, and concatenated with the upsampled decoder state:
\begin{equation}
\begin{aligned}
    Z_l
    &=
    [\widetilde{D}_l;
    \mathcal{B}_K(\widehat{S}_l);
    \mathcal{B}_K(\widehat{G}_l)],\\
    D_l
    &=
    \mathrm{DoubleConv}(Z_l)
    \in
    \mathbb{R}^{(BK)\times C_l\times H_l\times W_l}.
\end{aligned}
\end{equation}
where $\mathcal{B}_K(\cdot)$ denotes broadcasting over the $K$ categories and folding the batch and category dimensions, and $\mathrm{DoubleConv}$ consists of two $3\times3$ convolutions, each followed by group normalization and ReLU.

Progressive decoding is performed at $l=2$ and $l=3$ using $\{S_2,G_2\}$ and $\{S_3,G_3\}$, respectively. The first stage produces $D_2$, which is subsequently upsampled and passed to the second stage to obtain $D_3$. A prediction head then maps each category-specific feature in $D_3$ to a single logit map:
\begin{equation}
    \widehat{Y}^{\mathrm{dec}}
    =
    \mathrm{reshape}_{B,K}
    \left(
    \mathrm{Head}(D_3)
    \right)
    \in
    \mathbb{R}^{B\times K\times H_3\times W_3}.
\end{equation}
%

Each decoding stage upsamples its input by a factor of two through transposed convolution, yielding $H_3=4H_0$ and $W_3=4W_0$ after the two stages. Since the resulting resolution remains lower than the input resolution, we resize $\widehat{Y}^{\mathrm{dec}}$ to $H\times W$ using parameter-free bilinear interpolation, producing the final pixel-level logits $\widehat{Y}\in\mathbb{R}^{B\times K\times H\times W}$.




\subsection{Training Objective}
\label{sec:loss}

GeoSeg-OV is optimized using a per-pixel binary cross-entropy loss without auxiliary supervision.

During training, the auxiliary VFM remains fully frozen. For CLIP, we fine-tune only the query and value projections in its attention layers to enable limited adaptation to remote sensing imagery, while keeping all other parameters fixed. The remaining trainable components include the cost embedding and aggregation layers, the structural projection layers in SGA, the CAD refinement modules, and the decoder. This selective optimization strategy preserves most of the pretrained knowledge in CLIP and the auxiliary VFM while allowing the newly introduced aggregation and decoding components to adapt to remote sensing segmentation.

\section{Experiments}

\subsection{Experimental Setup}

\subsubsection{Benchmark Construction}

Existing evaluations of open-vocabulary remote sensing segmentation typically test on one or two datasets with similar acquisition conditions, which limits the assessment of cross-dataset generalization. To address this, we construct a global HRLC benchmark comprising seven datasets that span diverse geographic regions, acquisition platforms, and spatial resolutions as shown in Fig.~\ref{fig:benchmark}. Two datasets serve as training sources; the remaining five are used for cross-dataset evaluation without target-domain adaptation. When one training dataset is selected, the other is included in the evaluation set, yielding six evaluation datasets per training setting.

\noindent\textbf{Training datasets.}
\textit{FLAIR}~\citep{garioud2023flair} is a large-scale French land cover dataset acquired by the National Institute of Geographic and Forest Information (IGN) from aerial platforms. It covers over 77 cities across metropolitan France at 0.2\,m/pixel, annotated with 12 categories: building, pervious surface, impervious surface, bare soil, water, coniferous, deciduous, brushwood, vineyard, herbaceous vegetation, agricultural land, and plowed land.

\textit{OpenEarthMap}~\citep{xia2023openearthmap} is a global land cover dataset collected from both satellite and aerial platforms, spanning 44 cities across 6 continents at 0.25--0.5\,m/pixel. It is annotated with 8 categories: bareland, rangeland, developed space, road, tree, water, agriculture land, and building.

\noindent\textbf{Evaluation datasets.}
\textit{LoveDA}~\citep{wang2021loveda} covers urban and rural scenes from three Chinese cities (Nanjing, Changzhou, Wuhan) acquired via Google Earth at 0.3\,m/pixel, with 7 categories: background, building, road, water, barren, forest, and agricultural.

\textit{EarthMiss}~\citep{zhou2026remote} comprises 3{,}355 Maxar WorldView-2 satellite images covering 13 cities across 5 continents at 0.6\,m/pixel, annotated with 8 categories: building, road, water, bareland, forest, farmland, playground, and background.

\textit{DeepGlobe}~\citep{demir2018deepglobe} is a satellite land cover dataset from DigitalGlobe Vivid+ imagery at 0.5\,m/pixel, with 6 categories: urban, agriculture, rangeland, forest, water, and barren.

\textit{Potsdam}~\citep{rottensteiner2014results} is captured by an UltraCamXp aerial camera over Potsdam, Germany at 0.05\,m/pixel, with 6 categories: impervious surface, building, low vegetation, tree, car, and clutter.

\textit{Vaihingen}~\citep{rottensteiner2014results} is acquired by an Intergraph/ZI DMC aerial camera over Vaihingen, Germany at 0.09\,m/pixel, sharing the same 6 categories as Potsdam.

Taken together, the benchmark exposes models to substantial cross-dataset shifts in geographic coverage, acquisition conditions, spatial resolution, and label taxonomy. Although some datasets share broad platform types or overlapping resolution ranges, each evaluation dataset differs from the selected training source in its overall combination of these factors and is evaluated without target-domain adaptation.

\subsubsection{Category and Prompt Protocol}

All methods use the original category names from each dataset as text inputs, without any cross-dataset name mapping, synonym substitution, or prompt optimization.
In our experiments, we adopt a fixed set of prompt templates (e.g., ``\textbf{\textit{A photo of a \{class\} in the scene.}}'') shared across all datasets and methods.

Different datasets may define semantically related but lexically distinct categories at different levels of granularity. For example, ``impervious surface'' (FLAIR/Potsdam), ``developed space'' (OpenEarthMap), and ``urban'' (DeepGlobe) refer to overlapping concepts yet use different names. We deliberately treat these as separate categories rather than mapping them to a unified label, because the ability to distinguish and correctly assign such fine-grained, dataset-specific vocabulary is precisely what open-vocabulary segmentation should evaluate. A category is considered ``seen'' if its name or a near-identical variant (e.g., ``bareland'' vs.\ ``bare soil'') appears in the training vocabulary; semantically related but lexically distinct labels (e.g., ``impervious surface'' vs.\ ``developed space'' vs.\ ``urban'') are treated as unseen.

All annotated categories in each evaluation dataset, including background and clutter classes, participate in metric computation without exclusion.

\subsubsection{Evaluation Protocol}

We adopt a cross-dataset evaluation protocol without target-domain adaptation: models are trained on one source dataset and evaluated on all other datasets without fine-tuning or adaptation, directly measuring open-vocabulary generalization under domain shift.

We report three standard metrics. Let $TP_k$, $FP_k$, and $FN_k$ denote the true positives, false positives, and false negatives for category $k$, and let $N = \sum_{k=1}^{K}(TP_k + FN_k)$ be the total number of labeled pixels.

Mean Intersection over Union (mIoU) measures the average per-category overlap:
\begin{equation}
    \mathrm{mIoU} = \frac{1}{K} \sum_{k=1}^{K} \frac{TP_k}{TP_k + FP_k + FN_k}.
\end{equation}

Frequency-weighted IoU (fwIoU) weights each category by its pixel frequency:
\begin{equation}
    \mathrm{fwIoU} = \frac{1}{N} \sum_{k=1}^{K} (TP_k + FN_k) \cdot \frac{TP_k}{TP_k + FP_k + FN_k}.
\end{equation}

Mean pixel accuracy (mACC) computes the average per-category recall:
\begin{equation}
    \mathrm{mACC} = \frac{1}{K} \sum_{k=1}^{K} \frac{TP_k}{TP_k + FN_k}.
\end{equation}

The primary comparison metric is the average mIoU across all evaluation datasets, reflecting overall cross-dataset generalization capability.

\subsubsection{Compared Methods}

We compare GeoSeg-OV against eight representative methods spanning training-free, general trainable, and remote sensing open-vocabulary segmentation approaches:

\begin{itemize}
    \item \textbf{ClearCLIP}~\citep{lan2024clearclip} (ECCV'24) enhances CLIP for training-free open-vocabulary segmentation by removing noisy self-attention features and retaining only the residual pathway, thereby improving spatial localization without any task-specific training.
    \item \textbf{SegEarth-OV}~\citep{li2025segearth} (CVPR'25) is a training-free framework that adapts vision--language models for remote sensing segmentation by exploiting CLIP visual features with tailored spatial processing for overhead imagery.
    \item \textbf{SAN}~\citep{xu2023side} (CVPR'23) introduces side adapter networks to extract multi-scale features from a frozen CLIP encoder, enabling dense prediction while preserving open-vocabulary alignment.
    \item \textbf{SED}~\citep{xie2024sed} (CVPR'24) proposes a simple encoder-decoder architecture that directly bridges CLIP features to dense segmentation without relying on external mask generators.
    \item \textbf{CAT-Seg}~\citep{cho2024cat} (CVPR'24) formulates open-vocabulary segmentation as cost aggregation, constructing a vision--language cost volume from CLIP features and refining it through Swin-based transformer layers.
    \item \textbf{FGA-Seg}~\citep{li2025fgaseg} (ArXiv'25) introduces pixel-level visual--text alignment and convolution-based local similarity maps to refine boundary details within the cost aggregation framework.
    \item \textbf{OVRS}~\citep{cao2025open} (TGRS'25) extends CAT-Seg with multi-rotation cost construction, encoding the input under four orientations to address rotation ambiguity in overhead imagery.
    \item \textbf{GSNet}~\citep{ye2025towards} (AAAI'25) pairs CLIP with a remote-sensing-pretrained DINO backbone (RSIB) in a dual-stream architecture, fusing auxiliary features into the cost volume through query-guided correlation.
    \item \textbf{RSKT-Seg}~\citep{li2026exploring} (AAAI'26) integrates multi-direction cost aggregation with an efficient fusion transformer and remote sensing knowledge transfer from domain-adapted encoders including RemoteCLIP and DINO~\citep{caron2021emerging}.
\end{itemize}

ClearCLIP and SegEarth-OV are training-free methods that require no task-specific optimization, serving as references for the zero-shot capability of vision--language models. SAN, SED, and CAT-Seg are general-purpose trainable OVSS methods designed for natural images; CAT-Seg additionally serves as the baseline upon which all remote sensing methods in this comparison are built. OVRS, GSNet, and RSKT-Seg represent the current SOTA methods for open-vocabulary remote sensing segmentation. For fair comparison, all cost-aggregation-based methods (CAT-Seg, OVRS, GSNet, RSKT-Seg, and GeoSeg-OV) share the same CLIP ViT-B/16 backbone and are trained with identical data splits, iteration counts, and batch sizes.

\subsubsection{Implementation Details}
We adopt CLIP ViT-B/16 as the vision--language backbone. GeoSeg-OV is configured with $L=2$ aggregation layers, $N_h=4$ attention heads, hidden dimension $C=128$, window size $W=12$, and input resolution $384 \times 384$. We use the AdamW optimizer with base learning rate $2 \times 10^{-4}$, cosine schedule, and a $0.01\times$ multiplier for CLIP attention parameters. All trainable methods are trained for 30{,}000 iterations with batch size 4 on a single NVIDIA RTX 4090 GPU.

For the auxiliary structural encoder, we evaluate three vision foundation models: DINOv2 ViT-B/14~\citep{oquab2023dinov2}, SAM 2.1 Hiera Base Plus~\citep{ravi2024sam}, and Depth Anything V2 ViT-B/14~\citep{yang2024depth}. All auxiliary encoders remain entirely frozen during training. We adopt Depth Anything V2 as the default configuration; the comparison among auxiliary VFMs is presented in Table~\ref{tab:vfm-unified-heat}.

\begin{table*}[!t]
\centering
\scriptsize
\definecolor{LTabBlue}{RGB}{229,243,255}
\caption{Cross-dataset evaluation results (\%) on the Global \textbf{HRLC} benchmark. Evaluation metrics include mIoU, fwIoU, and mACC. The best results are shown in \textbf{bold}.}
\label{tab:main-results}
\resizebox{\textwidth}{!}{%
  \setlength{\tabcolsep}{1.5pt}%
  \renewcommand{\arraystretch}{1.08}%
  \begin{tabular}{@{}l|*{7}{ccc}|ccc@{}}
  \toprule
  \multirow{2}{*}{\textbf{Method}}
   & \multicolumn{3}{c}{\textbf{LoveDA}} & \multicolumn{3}{c}{\textbf{EarthMiss}}
   & \multicolumn{3}{c}{\textbf{DeepGlobe}} & \multicolumn{3}{c}{\textbf{Potsdam}}
   & \multicolumn{3}{c}{\textbf{Vaihingen}} & \multicolumn{3}{c}{\textbf{FLAIR}}
   & \multicolumn{3}{c|}{\textbf{OpenEarthMap}}
   & \multicolumn{3}{c}{\textbf{Average}} \\
  \cmidrule(r){2-4}\cmidrule(lr){5-7}\cmidrule(lr){8-10}\cmidrule(lr){11-13}%
  \cmidrule(lr){14-16}\cmidrule(lr){17-19}\cmidrule(lr){20-22}\cmidrule(l){23-25}
   & mIoU & fwIoU & mACC & mIoU & fwIoU & mACC & mIoU & fwIoU & mACC
   & mIoU & fwIoU & mACC & mIoU & fwIoU & mACC & mIoU & fwIoU & mACC
   & mIoU & fwIoU & mACC & mIoU & fwIoU & mACC \\
  \midrule
  \multicolumn{25}{c}{\textbf{\textit{Training-Free based methods}}} \\
  \midrule
  ClearCLIP\,{\textcolor{black!62}{\fontsize{3.85}{4.2}\selectfont ECCV'24}} &
  22.3 & — & 44.4 & 22.6 & — & 40.5 & 34.0 & — & 47.7 & 27.4 & — & 40.4 & 15.6 & — & 28.6 & 18.9 & — & 37.2 & 26.7 & — & 44.1 & 23.9 & — & 40.4 \\
  SegEarth-OV\,{\textcolor{black!62}{\fontsize{3.85}{4.2}\selectfont CVPR'25}} &
  34.1 & — & 62.6 & 33.7 & — & 57.6 & 42.9 & — & 62.3 & 44.9 & — & 63.7 & 39.0 & — & 63.0 & 20.4 & — & 40.7 & 37.9 & — & 60.9 & 36.1 & — & 58.7 \\
  \midrule
  \multicolumn{25}{c}{\textbf{\textit{FLAIR as the training dataset}}} \\
  \midrule
  SAN\,{\textcolor{black!62}{\fontsize{3.85}{4.2}\selectfont CVPR'23}} &
  27.2 & 25.5 & 54.6 & 24.8 & 25.9 & 48.0 & 30.9 & 41.6 & 51.0 & 27.4 & 44.8 & 44.4 & 21.8 & 32.7 & 34.7 & — & — & — & 29.1 & 30.7 & 48.2 & 26.9 & 33.5 & 46.8 \\
  SED\,{\textcolor{black!62}{\fontsize{3.85}{4.2}\selectfont CVPR'24}} &
  33.4 & 30.0 & 62.5 & 31.3 & 29.3 & 53.6 & 35.8 & 49.4 & 57.0 & 38.4 & 48.3 & 51.7 & 40.5 & 57.3 & 54.3 & — & — & — & 36.0 & 36.2 & 57.7 & 35.9 & 41.8 & 56.1 \\
  CAT-Seg\,{\textcolor{black!62}{\fontsize{3.85}{4.2}\selectfont CVPR'24}} &
  38.0 & 33.1 & 65.9 & 37.1 & 35.6 & 58.4 & 46.5 & 60.5 & 64.0 & 36.2 & 52.9 & 47.4 & 38.3 & 54.1 & 49.2 & — & — & — & 39.5 & 38.2 & 64.4 & 39.3 & 45.7 & 58.2 \\
  FGA-Seg\,{\textcolor{black!62}{\fontsize{3.85}{4.2}\selectfont ArXiv'25}} &
  38.0 & 32.8 & 67.1 & 37.3 & 34.8 & 60.4 & 45.3 & 58.4 & 65.0 & 38.2 & 53.7 & 49.5 & 39.4 & 56.1 & 49.8 & — & — & — & 40.3 & 39.2 & 64.4 & 39.7 & 45.8 & 59.4 \\
  OVRS\,{\textcolor{black!62}{\fontsize{3.85}{4.2}\selectfont TGRS'25}} &
  39.4 & 35.2 & 66.4 & 37.8 & 37.2 & 60.9 & \textbf{48.1} & \textbf{62.1} & 65.3 & 43.6 & 56.3 & 54.5 & 40.7 & 57.2 & 52.9 & — & — & — & 40.6 & 39.7 & 64.9 & 41.7 & 47.9 & 60.8 \\
  GSNet\,{\textcolor{black!62}{\fontsize{3.85}{4.2}\selectfont AAAI'25}} &
  37.1 & 32.5 & 65.6 & 36.5 & 34.3 & 58.6 & 44.7 & 59.0 & 64.5 & 36.1 & 52.6 & 47.3 & 40.4 & 57.4 & 50.6 & — & — & — & 38.6 & 36.9 & 64.6 & 38.9 & 45.5 & 58.5 \\
  RSKT-Seg\,{\textcolor{black!62}{\fontsize{3.85}{4.2}\selectfont AAAI'26}} &
  39.3 & 33.4 & \textbf{68.1} & 37.4 & 34.5 & 61.1 & 46.3 & 60.0 & \textbf{65.7} & 41.2 & 54.6 & 52.3 & 41.9 & 59.0 & 51.7 & — & — & — & 40.2 & 38.9 & 65.5 & 41.1 & 46.7 & 60.7 \\
  \rowcolor{LTabBlue}%
  \textbf{GeoSeg-OV (Ours)} &
  \textbf{40.7} & \textbf{38.0} & 66.4 &
  \textbf{40.5} & \textbf{42.0} & \textbf{62.6} &
  46.9 & 60.9 & 65.2 &
  \textbf{49.3} & \textbf{60.0} & \textbf{60.5} &
  \textbf{43.8} & \textbf{61.6} & \textbf{55.7} &
  — & — & — &
  \textbf{44.0} & \textbf{44.0} & \textbf{66.2} &
  \textbf{44.2} & \textbf{51.1} & \textbf{62.8} \\
  \midrule
  \multicolumn{25}{c}{\textbf{\textit{OpenEarthMap as the training dataset}}} \\
  \midrule
  SAN\,{\textcolor{black!62}{\fontsize{3.85}{4.2}\selectfont CVPR'23}} &
  30.2 & 29.0 & 54.9 & 29.2 & 35.8 & 53.0 & 35.3 & 51.6 & 47.3 & 25.4 & 47.3 & 46.2 & 22.1 & 31.6 & 35.8 & 22.4 & 27.9 & 40.8 & — & — & — & 27.4 & 37.2 & 46.3 \\
  SED\,{\textcolor{black!62}{\fontsize{3.85}{4.2}\selectfont CVPR'24}} &
  40.9 & 36.3 & 68.1 & 41.8 & 44.8 & 63.7 & 36.5 & 54.9 & 48.6 & 38.6 & 48.3 & 53.6 & 32.9 & 46.4 & 50.0 & 21.9 & 24.3 & 37.1 & — & — & — & 35.4 & 42.5 & 53.5 \\
  CAT-Seg\,{\textcolor{black!62}{\fontsize{3.85}{4.2}\selectfont CVPR'24}} &
  40.8 & 33.9 & 68.5 & 40.7 & 38.4 & 65.7 & 43.9 & 59.4 & 56.6 & 38.8 & 54.7 & 49.9 & 35.5 & 53.0 & 46.0 & 26.5 & 29.3 & 38.2 & — & — & — & 37.7 & 44.8 & 54.2 \\
  FGA-Seg\,{\textcolor{black!62}{\fontsize{3.85}{4.2}\selectfont ArXiv'25}} &
  40.3 & 33.1 & 68.4 & 39.3 & 36.1 & 65.1 & 40.6 & 57.2 & 52.8 & 40.1 & 54.3 & 55.6 & 34.9 & 50.3 & 49.3 & 27.2 & 30.6 & 39.9 & — & — & — & 37.1 & 43.6 & 55.2 \\
  OVRS\,{\textcolor{black!62}{\fontsize{3.85}{4.2}\selectfont TGRS'25}} &
  41.3 & 34.9 & 68.5 & 42.0 & 40.9 & 66.1 & 44.6 & 59.8 & 57.6 & 40.3 & 54.2 & 55.1 & 37.3 & 53.2 & 51.1 & 27.6 & 29.9 & 39.3 & — & — & — & 38.9 & 45.5 & 56.3 \\
  GSNet\,{\textcolor{black!62}{\fontsize{3.85}{4.2}\selectfont AAAI'25}} &
  41.3 & 34.4 & 68.5 & 40.8 & 39.7 & 65.3 & 43.4 & 59.1 & 56.5 & 40.3 & 55.8 & 52.5 & 36.1 & 53.4 & 47.2 & 26.5 & 30.1 & 38.4 & — & — & — & 38.1 & 45.4 & 54.7 \\
  RSKT-Seg\,{\textcolor{black!62}{\fontsize{3.85}{4.2}\selectfont AAAI'26}} &
  41.3 & 34.3 & 68.1 & 42.6 & 43.4 & 66.5 & 42.4 & 58.4 & 56.3 & 42.4 & 57.5 & 55.1 & 36.8 & 53.0 & 49.1 & 28.2 & 30.8 & 40.3 & — & — & — & 38.9 & 46.2 & 55.9 \\
  \rowcolor{LTabBlue}%
  \textbf{GeoSeg-OV (Ours)} &
  \textbf{42.6} & \textbf{37.1} & \textbf{68.8} &
  \textbf{45.2} & \textbf{46.4} & \textbf{68.2} &
  \textbf{44.7} & \textbf{59.9} & \textbf{57.9} &
  \textbf{47.7} & \textbf{59.5} & \textbf{62.1} &
  \textbf{39.9} & \textbf{56.7} & \textbf{51.7} &
  \textbf{29.2} & \textbf{30.9} & \textbf{41.0} &
  — & — & — &
  \textbf{41.6} & \textbf{48.4} & \textbf{58.3} \\
  \bottomrule
  \end{tabular}%
}
\end{table*}

\subsection{Main Results}

Table~\ref{tab:main-results} presents the cross-dataset evaluation results on the global HRLC benchmark. We report results under two training settings (FLAIR and OpenEarthMap) and additionally include two training-free methods for reference.

\noindent\textbf{Overall performance.}
GeoSeg-OV achieves the best average performance across all three metrics under both training settings: 44.2/41.6 mIoU, 51.1/48.4 fwIoU, and 62.8/58.3 mACC. Compared with the best-performing prior trainable methods, namely OVRS at 41.7 average mIoU under FLAIR and both OVRS and RSKT-Seg at 38.9 under OpenEarthMap, GeoSeg-OV improves by +2.5 and +2.7 average mIoU, respectively. Relative to the CAT-Seg baseline from which all remote sensing methods derive, the improvement reaches +4.9 and +3.9. Among training-free methods, SegEarth-OV achieves 36.1 average mIoU without any task-specific training. GeoSeg-OV consistently outperforms SegEarth-OV on every individual evaluation dataset under both training settings.

\noindent\textbf{Per-dataset analysis.}
GeoSeg-OV ranks first on 5 out of 6 evaluation datasets under FLAIR training and on all 6 under OpenEarthMap training. The gains are not uniform but correlate with the severity of domain shift.

On Potsdam, the resolution gap from the training data is the largest in our benchmark. GeoSeg-OV achieves 49.3/47.7 mIoU, outperforming the second-best method by +5.7/+5.3. At this resolution, fine-grained details such as roof textures, inter-building shadows, and individual vehicles create substantial appearance discrepancy from moderate-resolution training imagery. Structural cues such as object extent and boundary discontinuity remain stable despite the resolution mismatch, explaining the large margin.

On EarthMiss, GeoSeg-OV achieves 40.5/45.2 mIoU, improving by +2.7/+2.6 over the second-best methods. The geographic diversity of this dataset introduces wide cross-regional appearance variation, where structural relationships transfer more reliably than appearance statistics.

On DeepGlobe, GeoSeg-OV achieves 46.9/44.7 mIoU. Under the FLAIR setting, OVRS slightly outperforms our GeoSeg-OV (48.1 vs.\ 46.9), while under the OpenEarthMap setting GeoSeg-OV leads. DeepGlobe defines coarse-grained categories at moderate resolution where multi-rotation matching already provides strong signals, leaving less room for structural guidance to contribute additional gains.

\noindent\textbf{Comparison with AVTM methods.}
GSNet and RSKT-Seg both incorporate auxiliary foundation models (DINO, RemoteCLIP) but couple them with cost-map construction as additional matching evidence, achieving 38.9/38.1 and 41.1/38.9 average mIoU, respectively. GeoSeg-OV, using a single frozen encoder purely as structural priors outside the matching space, achieves 44.2/41.6. The consistent gap (+3.5--5.3 over GSNet; +2.7--3.1 over RSKT-Seg) across both settings indicates that decoupling auxiliary features from the matching space and redirecting them toward cost aggregation guidance is more effective for cross-dataset generalization.

\begin{figure*}[t]
    \centering
    \includegraphics[width=\textwidth]{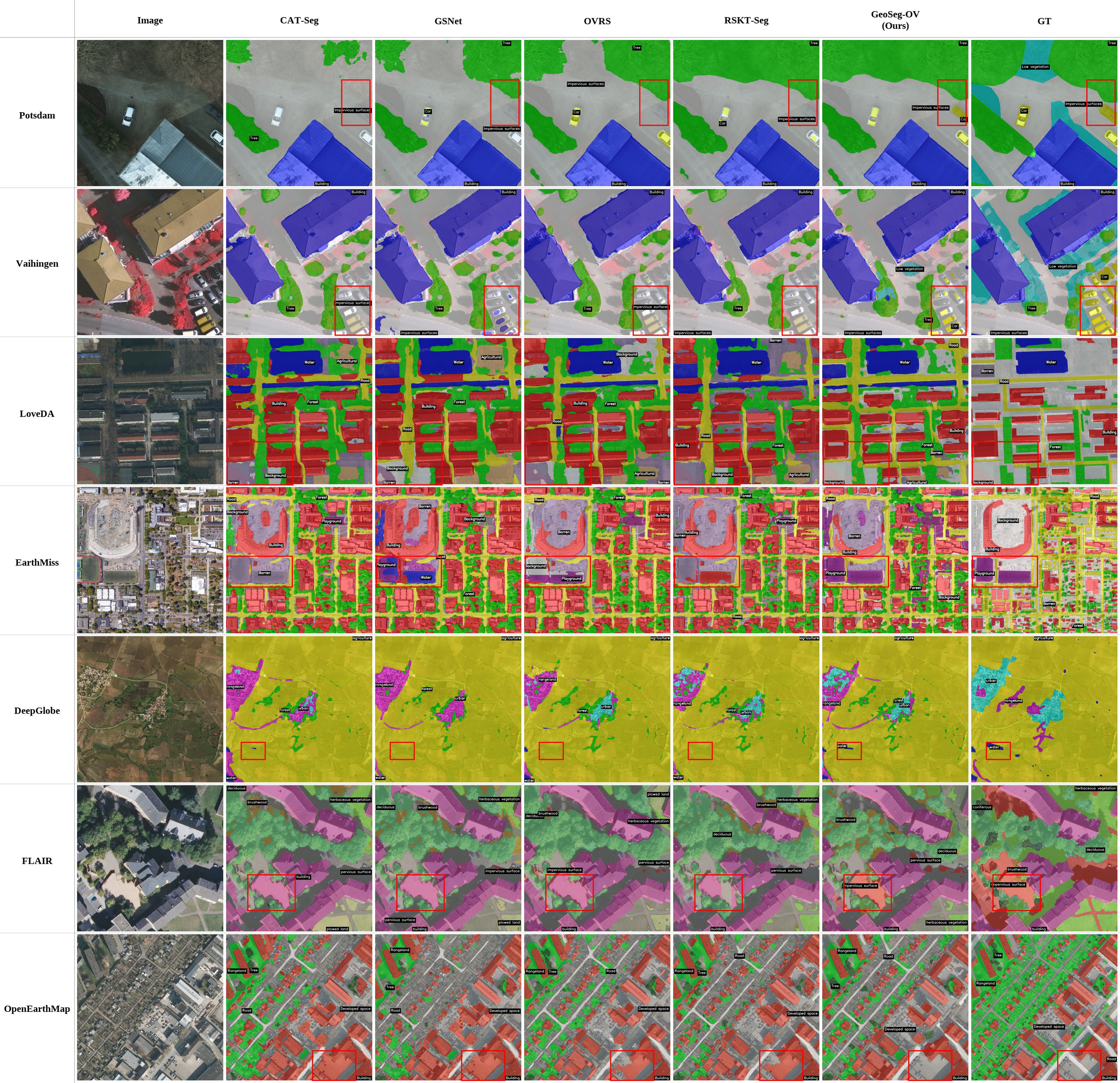}
    \caption{Qualitative comparison of segmentation results on cross-dataset evaluation samples. From left to right: input image, CAT-Seg, GSNet, OVRS, RSKT-Seg, GeoSeg-OV (Ours), and ground truth. Rows from top to bottom correspond to Potsdam, Vaihingen, LoveDA, EarthMiss, DeepGlobe, FLAIR, and OpenEarthMap.}
    \label{fig:qualitative}
\end{figure*}

\subsection{Qualitative Analysis}

Fig.~\ref{fig:qualitative} visualizes segmentation predictions from representative methods on cross-dataset evaluation scenes. Several patterns recur across datasets.

\noindent\textbf{Boundary preservation under resolution mismatch (Potsdam, Vaihingen).}
At very high resolution (0.05--0.09\,m), fine-grained details such as building shadows and roof textures create strong appearance ambiguity. Baseline methods tend to produce fragmented predictions or absorb small objects (e.g., vehicles) into surrounding impervious surfaces. GeoSeg-OV maintains sharper object boundaries and more compact predictions, indicating that structural priors help distinguish adjacent regions even when appearance cues are unreliable.

\noindent\textbf{Region coherence under geographic variation (LoveDA, EarthMiss, OpenEarthMap).}
In dense urban scenes, baseline methods often fragment building regions or allow predictions to bleed into adjacent surfaces. For spatially extensive categories (e.g., background, playground, developed space), cross-dataset appearance variation further confuses matching-based methods. GeoSeg-OV produces more coherent region extents and cleaner transitions between land cover types, reflecting the benefit of aggregation guided by region continuity.

\noindent\textbf{Disambiguation under appearance confusion (DeepGlobe, FLAIR).}
Water regions in DeepGlobe exhibit greenish tones visually similar to agricultural land, and FLAIR includes pervious and impervious surfaces with highly overlapping color distributions. Baseline methods produce mixed or confused predictions at these category boundaries. GeoSeg-OV generates more spatially consistent regions with cleaner boundary delineation, suggesting that structural guidance provides discriminative cues at surface transitions where appearance alone is insufficient.

\newcommand{\yes}{\checkmark}
\newcommand{\no}{}

\begin{table}[!t]
\centering
\caption{Component analysis of GeoSeg-OV. We evaluate the individual and combined effects of SGA, CAD, and multi-rotation encoding (Rot) on top of the baseline.}
\label{tab:ablation_component}

\scriptsize
\setlength{\tabcolsep}{2.2pt}
\renewcommand{\arraystretch}{1.05}

\resizebox{\columnwidth}{!}{%
\begin{tabular}{@{}cccc||ccc|ccc@{}}
\hline\hline
\multicolumn{4}{c||}{\textbf{Components}} &
\multicolumn{3}{c|}{\textbf{FLAIR}} &
\multicolumn{3}{c}{\textbf{OpenEarthMap}} \\
Baseline & Rot & SGA & CAD &
mIoU & fwIoU & mACC & mIoU & fwIoU & mACC \\
\hline\hline
\yes & \no & \no & \no &
39.3 & 45.8 & 58.2 & 37.7 & 44.8 & 54.2 \\
\yes & \no & \yes & \no &
41.8 & 48.2 & 60.8 & 39.8 & 46.8 & 56.7 \\
\yes & \no & \yes & \yes &
42.9 & 49.2 & 62.1 & 40.4 & 47.0 & 57.7 \\
\yes & \yes & \no & \no &
40.9 & 47.1 & 59.8 & 39.2 & 46.3 & 56.3 \\
\yes & \yes & \yes & \no &
42.8 & 49.7 & 61.3 & 40.6 & 47.9 & 57.4 \\
\cellcolor{blue!10}\yes & \cellcolor{blue!10}\yes & \cellcolor{blue!10}\yes & \cellcolor{blue!10}\yes &
\cellcolor{blue!10}\textbf{44.2} & \cellcolor{blue!10}\textbf{51.1} & \cellcolor{blue!10}\textbf{62.8} &
\cellcolor{blue!10}\textbf{41.6} & \cellcolor{blue!10}\textbf{48.4} & \cellcolor{blue!10}\textbf{58.3} \\
\hline\hline
\end{tabular}%
}
\end{table}

\subsection{Ablation Studies}

\subsubsection{Component Analysis}

Table~\ref{tab:ablation_component} presents an ablation study that progressively introduces the proposed components on top of the CAT-Seg baseline under both training settings.

\noindent\textbf{Combined effect of SGA and CAD.}
The two proposed modules together improve the baseline by +3.6/+2.7 mIoU without any rotation augmentation (42.9/40.4 vs.\ 39.3/37.7). This variant already surpasses all compared methods in Table~\ref{tab:main-results}, including OVRS (41.7/38.9), which employs multi-rotation encoding, and RSKT-Seg (41.1/38.9), which incorporates multiple auxiliary encoders. This confirms that the core contribution of GeoSeg-OV, using auxiliary features as structural priors rather than as additional matching evidence, is the primary source of improvement, independent of rotation augmentation.

\noindent\textbf{Role of SGA.}
SGA alone improves the baseline by +2.5/+2.1 mIoU (41.8/39.8 vs.\ 39.3/37.7), constituting the largest single-component gain. This is expected, as SGA introduces structural priors into the core aggregation operation that determines how matching evidence is exchanged across positions, producing more coherent cost representations for all subsequent stages. By injecting a structure-aware pairwise bias, SGA supplies explicit spatial-structural cues that content based attention alone captures only implicitly.

\noindent\textbf{Role of CAD.}
Adding CAD on top of SGA yields a further +1.1/+0.6 mIoU improvement (42.9/40.4 vs.\ 41.8/39.8). Although numerically smaller, CAD operates at a different stage: it refines the guidance features received by the decoder according to the current cost state. Without CAD, semantic and structural guidance are applied uniformly regardless of the segmentation state at each decoding scale, whereas CAD introduces cost-conditioned adaptivity that allows the decoder to emphasize guidance relevant to the current prediction.

\noindent\textbf{Compatibility with multi-rotation encoding.}
Multi-rotation cost construction is employed for handling orientation ambiguity in overhead imagery. When combined with SGA and CAD, it provides an additional +1.3/+1.2 mIoU, bringing the full model to 44.2/41.6. This indicates that rotation-robust cost construction and structure-guided aggregation address different aspects of the geospatial gap. Notably, SGA and CAD account for the majority of the total improvement, contributing +3.6/+2.7 out of +4.9/+3.9.

\begin{figure*}[t]
    \centering
    \includegraphics[width=\textwidth]{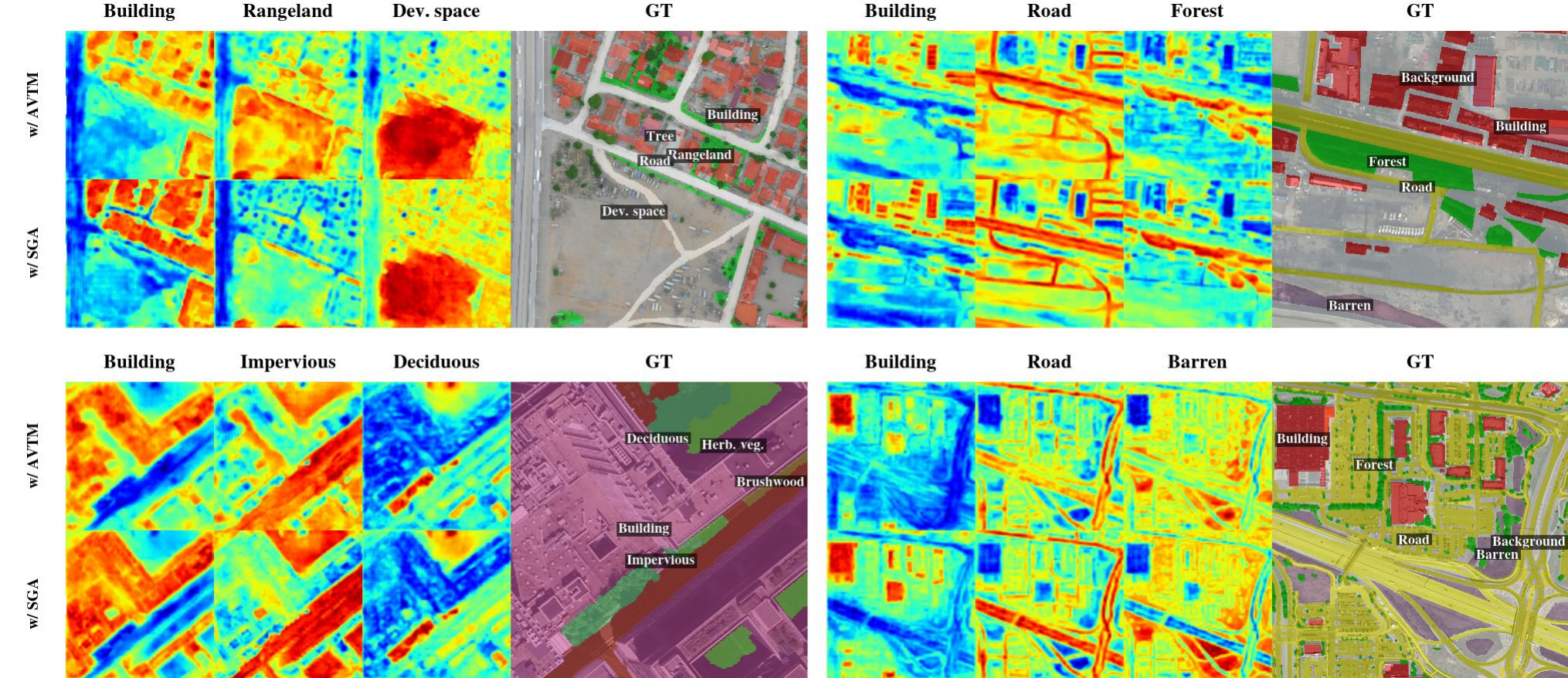}
    \caption{Visualization of class-wise aggregated cost maps under the AVTM paradigm and the proposed SGA paradigm. Each column shows the cost map for a specific category. SGA produces more coherent activations with cleaner spatial structure for categories distinguished by layout and boundary organization.}
    \label{fig:cost_map}
\end{figure*}

\begin{table}[!t]
  \centering
  \caption{Comparison of VFM utilization paradigms. \textit{AVTM} uses auxiliary features for cost volume construction; \textit{SGA} uses them to guide cost aggregation outside the matching space. Both are evaluated with two auxiliary encoders.}
  \label{tab:ablation_paradigm}
  \resizebox{\columnwidth}{!}{%
    \footnotesize
    \setlength{\tabcolsep}{4pt}
    \begin{tabular}{@{} c c || ccc | ccc @{}}
      \toprule
      \textbf{Paradigm}
      & \textbf{VFM}
      & \multicolumn{3}{c|}{\textbf{FLAIR}}
      & \multicolumn{3}{c}{\textbf{OpenEarthMap}} \\
      & & mIoU & fwIoU & mACC & mIoU & fwIoU & mACC \\
      \midrule
      \multirow{2}{*}{AVTM}
        & DA-V2 
        & 39.8 & 47.5 & 59.1 & 38.6 & 46.2 & 55.1 \\
      & RSIB-DINO      
        & 38.9 & 45.5 & 58.5 & 38.1 & 45.4 & 54.7 \\
      \midrule
      \multirow{2}{*}{SGA}
        & DA-V2
        & 41.8 & 48.2 & 60.8 & 39.8 & 46.8 & 56.7 \\
      & RSIB-DINO
        & 40.2 & 46.7 & 59.4 & 39.2 & 46.1 & 56.0 \\
      \midrule
      \multirow{2}{*}{$\Delta$}
        & DA-V2
        & \textcolor{green!50!black}{+2.0}
        & \textcolor{green!50!black}{+0.7}
        & \textcolor{green!50!black}{+1.7}
        & \textcolor{green!50!black}{+1.2}
        & \textcolor{green!50!black}{+0.6}
        & \textcolor{green!50!black}{+1.6} \\
      & RSIB-DINO
        & \textcolor{green!50!black}{+1.3}
        & \textcolor{green!50!black}{+1.2}
        & \textcolor{green!50!black}{+0.9}
        & \textcolor{green!50!black}{+1.1}
        & \textcolor{green!50!black}{+0.7}
        & \textcolor{green!50!black}{+1.3} \\
      \bottomrule
    \end{tabular}%
  }
\end{table}

\subsubsection{Effectiveness of Structure-Guided Aggregation}
\label{sec:sga_ablation}

We compare SGA with the AVTM paradigm used in existing methods~\citep{ye2025towards,li2026exploring}, where auxiliary features are matched with text embeddings to construct an additional cost map. Both variants use the same baseline and training configuration, with multi-rotation encoding and CAD disabled; their only difference lies in whether the auxiliary features are used for visual--text matching or structure-guided aggregation. We conduct this comparison using two auxiliary encoders, DA-V2 and RSIB-DINO, to evaluate the generality of SGA across different feature representations.

As shown in Table~\ref{tab:ablation_paradigm}, SGA consistently outperforms AVTM with both encoders. Compared with AVTM, SGA improves mIoU by +2.0/+1.2 with DA-V2 and +1.3/+1.1 with RSIB-DINO under the FLAIR/OpenEarthMap training settings. These controlled results demonstrate that using auxiliary features as structural guidance is more effective than treating them as additional matching evidence, regardless of the auxiliary encoder employed.

Fig.~\ref{fig:cost_map} further shows that SGA produces more coherent cost responses and sharper boundaries for categories with distinctive spatial structures, such as road, building, and developed space. In contrast, AVTM exhibits more diffuse activations across category boundaries, demonstrating the advantage of SGA in constraining the propagation of matching evidence to structurally coherent regions.

\begin{figure*}[t]
    \centering
    \includegraphics[width=\textwidth]{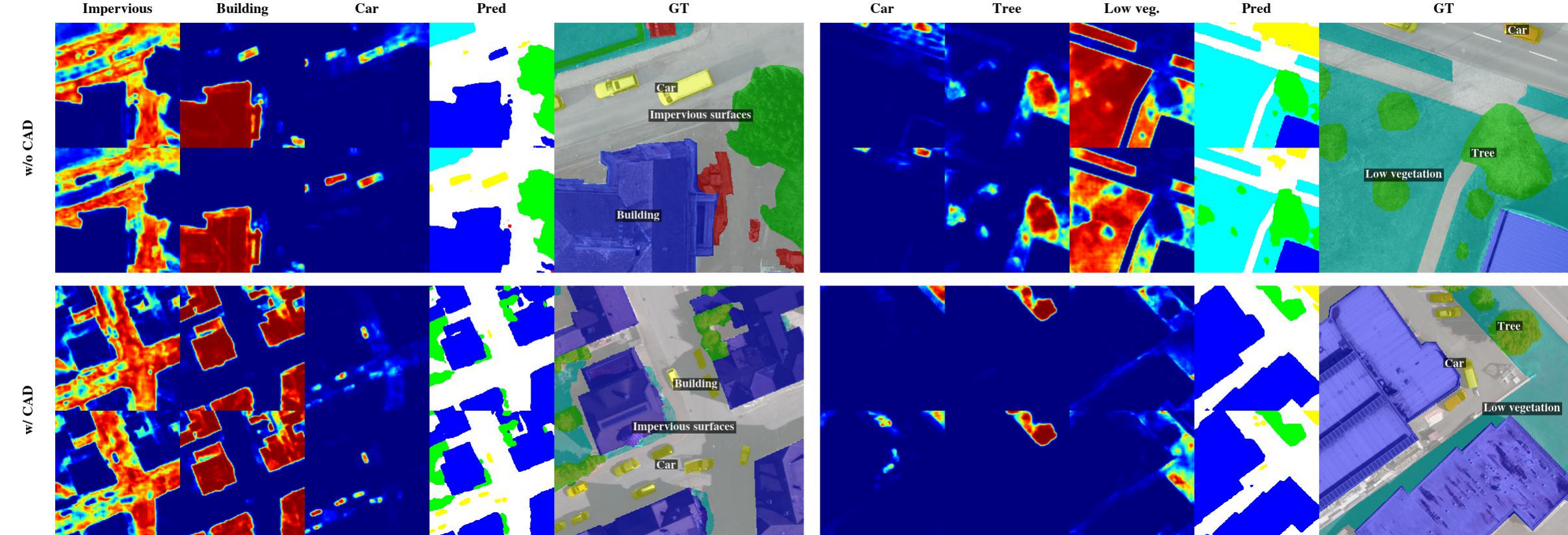}
    \caption{Per-category decoder logit maps with and without CAD on two high-resolution aerial scenes. Without CAD, logits for small objects are weak and fragmented while surrounding dominant categories spread over their locations; with CAD, small-object responses become localized and inter-category spillover is suppressed.}
    \label{fig:cad_logit}
\end{figure*}

\subsubsection{Effectiveness of Cost-Aware Decoding}
\label{sec:cad_vis}

Fig.~\ref{fig:cad_logit} visualizes per-category decoder logits with and without CAD on two high-resolution aerial scenes. Both scenes contain small objects located on or near visually similar surrounding surfaces, a challenging scenario where strong appearance overlap and shadow effects make these objects difficult to separate from dominant neighboring categories.

Without CAD, the decoder applies guidance uniformly: logits for dominant surrounding categories spread over small-object locations, while the small-object logits remain weak and spatially fragmented, causing them to be absorbed into neighboring predictions. With CAD, small-object responses become localized and the spillover from surrounding categories is suppressed. This behavior is consistent with the conditioning mechanism of CAD: at each decoding scale, the relevance gate is jointly determined by the pooled decoder context and the corresponding guidance features, allowing the refinement to adapt across spatial locations and channels.


\definecolor{heatlow}{RGB}{255, 252, 245}
\definecolor{heatmid}{RGB}{255, 228, 196}
\definecolor{heathigh}{RGB}{255, 194, 158}

\newcommand{\HLo}{\cellcolor{heatlow}}
\newcommand{\HMid}{\cellcolor{heatmid}}
\newcommand{\HHi}{\cellcolor{heathigh}}

\begin{table*}[!t]
\centering
\caption{Ablation on auxiliary vision foundation models. We evaluate DINO, SAM, and Depth as auxiliary structural guidance sources within the proposed \textbf{GeoSeg-OV} framework under both training settings.}
\label{tab:vfm-unified-heat}
\scriptsize
\resizebox{\textwidth}{!}{%
  \setlength{\tabcolsep}{1.4pt}%
  \renewcommand{\arraystretch}{1.08}%
  \begin{tabular}{@{}l|*{7}{ccc}|ccc@{}}
  \toprule
  \multicolumn{25}{c}{\textit{\textbf{FLAIR} as training dataset.}} \\
  \midrule
  \multirow{2}{*}{\textbf{VFM}}
   & \multicolumn{3}{c}{\textbf{LoveDA}} & \multicolumn{3}{c}{\textbf{EarthMiss}}
   & \multicolumn{3}{c}{\textbf{DeepGlobe}} & \multicolumn{3}{c}{\textbf{Potsdam}}
   & \multicolumn{3}{c}{\textbf{Vaihingen}} & \multicolumn{3}{c}{\textbf{FLAIR}}
   & \multicolumn{3}{c|}{\textbf{OpenEarthMap}} & \multicolumn{3}{c}{\textbf{Average}} \\
  \cmidrule(r){2-4}\cmidrule(lr){5-7}\cmidrule(lr){8-10}\cmidrule(lr){11-13}%
  \cmidrule(lr){14-16}\cmidrule(lr){17-19}\cmidrule(lr){20-22}\cmidrule(lr){23-25}
   & mIoU & fwIoU & mACC & mIoU & fwIoU & mACC & mIoU & fwIoU & mACC & mIoU & fwIoU & mACC
   & mIoU & fwIoU & mACC & mIoU & fwIoU & mACC & mIoU & fwIoU & mACC & mIoU & fwIoU & mACC \\
  \midrule
  \textbf{DINO} &
  \HHi\textbf{41.3} &
  \HHi\textbf{39.8} &
  \HLo 65.8 &
  \HHi\textbf{41.3} &
  \HHi\textbf{44.3} &
  \HLo 61.1 &
  \HLo 46.1 &
  \HLo 60.5 &
  \HLo 64.1 &
  \HLo 48.5 &
  \HMid 58.5 &
  \HLo 59.7 &
  \HLo 42.4 &
  \HMid 58.5 &
  \HLo 54.3 &
  --- & --- & --- &
  \HMid 43.4 &
  \HMid 42.9 &
  \HHi\textbf{66.3} &
  \HLo 43.8 &
  \HMid 50.7 &
  \HLo 61.9 \\
  \textbf{SAM} &
  \HLo 40.1 &
  \HLo 36.7 &
  \HHi\textbf{66.6} &
  \HLo 39.7 &
  \HLo 40.2 &
  \HHi\textbf{63.0} &
  \HHi\textbf{47.2} &
  \HHi\textbf{61.5} &
  \HHi\textbf{65.8} &
  \HHi\textbf{49.8} &
  \HLo 58.4 &
  \HHi\textbf{60.8} &
  \HHi\textbf{43.8} &
  \HLo 58.2 &
  \HHi\textbf{56.0} &
  --- & --- & --- &
  \HLo 42.7 &
  \HLo 42.3 &
  \HLo 65.9 &
  \HMid 43.9 &
  \HLo 49.6 &
  \HHi\textbf{63.0} \\
  \textbf{Depth} &
  \HMid 40.7 &
  \HMid 38.0 &
  \HMid 66.4 &
  \HMid 40.5 &
  \HMid 42.0 &
  \HMid 62.6 &
  \HMid 46.9 &
  \HMid 60.9 &
  \HMid 65.2 &
  \HMid 49.3 &
  \HHi\textbf{60.0} &
  \HMid 60.5 &
  \HMid 43.8 &
  \HHi\textbf{61.6} &
  \HMid 55.7 &
  --- & --- & --- &
  \HHi\textbf{44.0} &
  \HHi\textbf{44.0} &
  \HMid 66.2 &
  \HHi\textbf{44.2} &
  \HHi\textbf{51.1} &
  \HMid 62.8 \\
  \midrule
  \multicolumn{25}{c}{\textit{\textbf{OpenEarthMap} as training dataset.}} \\
  \midrule
  \multirow{2}{*}{\textbf{VFM}}
   & \multicolumn{3}{c}{\textbf{LoveDA}} & \multicolumn{3}{c}{\textbf{EarthMiss}}
   & \multicolumn{3}{c}{\textbf{DeepGlobe}} & \multicolumn{3}{c}{\textbf{Potsdam}}
   & \multicolumn{3}{c}{\textbf{Vaihingen}} & \multicolumn{3}{c}{\textbf{FLAIR}}
   & \multicolumn{3}{c|}{\textbf{OpenEarthMap}} & \multicolumn{3}{c}{\textbf{Average}} \\
  \cmidrule(r){2-4}\cmidrule(lr){5-7}\cmidrule(lr){8-10}\cmidrule(lr){11-13}%
  \cmidrule(lr){14-16}\cmidrule(lr){17-19}\cmidrule(lr){20-22}\cmidrule(lr){23-25}
   & mIoU & fwIoU & mACC & mIoU & fwIoU & mACC & mIoU & fwIoU & mACC & mIoU & fwIoU & mACC
   & mIoU & fwIoU & mACC & mIoU & fwIoU & mACC & mIoU & fwIoU & mACC & mIoU & fwIoU & mACC \\
  \midrule
  \textbf{DINO} &
  \HMid 42.4 &
  \HMid 37.0 &
  \HHi\textbf{68.8} &
  \HMid 44.8 &
  \HMid 46.2 &
  \HLo 67.5 &
  \HLo 43.5 &
  \HLo 59.1 &
  \HLo 56.8 &
  \HMid 46.4 &
  \HMid 59.3 &
  \HHi\textbf{62.1} &
  \HLo 38.6 &
  \HMid 55.3 &
  \HLo 51.3 &
  \HMid 29.1 &
  \HHi\textbf{30.9} &
  \HHi\textbf{41.3} &
  --- & --- & --- &
  \HLo 40.8 &
  \HMid 48.0 &
  \HLo 57.9 \\
  \textbf{SAM} &
  \HLo 41.8 &
  \HLo 35.8 &
  \HLo 68.7 &
  \HLo 43.4 &
  \HLo 43.0 &
  \HMid 67.9 &
  \HHi\textbf{45.0} &
  \HHi\textbf{59.9} &
  \HHi\textbf{58.3} &
  \HLo 46.3 &
  \HLo 56.8 &
  \HLo 61.7 &
  \HHi\textbf{42.2} &
  \HLo 54.7 &
  \HHi\textbf{57.2} &
  \HLo 28.6 &
  \HLo 30.5 &
  \HLo 40.6 &
  --- & --- & --- &
  \HMid 41.2 &
  \HLo 46.8 &
  \HHi\textbf{59.1} \\
  \textbf{Depth} &
  \HHi\textbf{42.6} &
  \HHi\textbf{37.1} &
  \HMid 68.8 &
  \HHi\textbf{45.2} &
  \HHi\textbf{46.4} &
  \HHi\textbf{68.2} &
  \HMid 44.7 &
  \HMid 59.9 &
  \HMid 57.9 &
  \HHi\textbf{47.7} &
  \HHi\textbf{59.5} &
  \HMid 62.1 &
  \HMid 39.9 &
  \HHi\textbf{56.7} &
  \HMid 51.7 &
  \HHi\textbf{29.2} &
  \HMid 30.9 &
  \HMid 41.0 &
  --- & --- & --- &
  \HHi\textbf{41.6} &
  \HHi\textbf{48.4} &
  \HMid 58.3 \\
  \bottomrule
  \end{tabular}%
}
\vspace{-0.4em}
\end{table*}

\begin{table*}[t]
\centering
\caption{Class-wise IoU (\%) for cross-domain open-vocabulary evaluation. \colorbox{blue!12}{Blue} headers denote seen classes (overlapping with training vocabulary); \colorbox{green!15}{Green} headers denote unseen classes (absent from training).}
\label{tab:classwise-flair-oem-combined}
\footnotesize
\setlength{\tabcolsep}{2.0pt}
\setlength{\doublerulesep}{0.45mm}
\renewcommand{\arraystretch}{1.07}
\resizebox{\textwidth}{!}{%
\begin{tabular}{@{}
  >{\raggedright\arraybackslash}p{2.45cm}|
  *{8}{c}|
  >{\centering\arraybackslash}m{0.78cm}
  >{\centering\arraybackslash}m{0.78cm}
  >{\centering\arraybackslash}m{0.78cm}
  ||
  *{12}{c}|
  >{\centering\arraybackslash}m{0.78cm}
  >{\centering\arraybackslash}m{0.78cm}
  >{\centering\arraybackslash}m{0.78cm}
@{}}
\toprule
\multirow{2}{*}[-2pt]{\textbf{Method}}
  & \multicolumn{11}{c|@{\hskip\doublerulesep}|}{\textbf{\textit{FLAIR $\rightarrow$ OpenEarthMap}}}
  & \multicolumn{15}{c@{}}{\textbf{\textit{OpenEarthMap $\rightarrow$ FLAIR}}} \\
\cmidrule(lr){2-9}\cmidrule(lr){10-12}\cmidrule(lr){13-24}\cmidrule(lr){25-27}
  & {\cellcolor{blue!12}Agri.}
  & {\cellcolor{blue!12}Bare.}
  & {\cellcolor{blue!12}Bldg.}
  & {\cellcolor{blue!12}Wat.}
  & {\cellcolor{green!15}Tree}
  & {\cellcolor{green!15}Dev.}
  & {\cellcolor{green!15}Rng.}
  & {\cellcolor{green!15}Rd.}
  & Seen & Unseen & All
  & {\cellcolor{blue!12}Agri.}
  & {\cellcolor{blue!12}B.\,soil}
  & {\cellcolor{blue!12}Bldg.}
  & {\cellcolor{blue!12}Wat.}
  & {\cellcolor{green!15}Dec.}
  & {\cellcolor{green!15}Br.}
  & {\cellcolor{green!15}Cf.}
  & {\cellcolor{green!15}Hb.}
  & {\cellcolor{green!15}Imp.}
  & {\cellcolor{green!15}Prv.}
  & {\cellcolor{green!15}Plw.}
  & {\cellcolor{green!15}Vin.}
  & Seen & Unseen & All \\
\midrule
SAN \textcolor{black!62}{\tiny CVPR'23} & 34.8 & 1.0 & 53.5 & 54.8 & 44.5 & 20.4 & 11.5 & 12.3 & 36.0 & 22.2 & 29.1 & 36.3 & 2.0 & 64.3 & 46.1 & 15.2 & 4.8 & 11.2 & 10.2 & 23.7 & 5.5 & 2.9 & \textbf{47.0} & 37.2 & 15.1 & 22.4 \\
SED \textcolor{black!62}{\tiny CVPR'24} & 58.6 & 2.8 & 63.1 & 60.0 & 58.9 & 8.5 & 10.1 & 26.3 & 46.1 & 26.0 & 36.0 & 37.4 & 0.5 & 75.4 & 57.7 & 24.8 & 5.0 & \textbf{13.1} & 0.0 & \textbf{38.9} & 7.1 & 2.9 & 0.2 & 42.8 & 11.5 & 21.9 \\
CAT-Seg \textcolor{black!62}{\tiny CVPR'24} & 60.7 & 9.3 & 65.2 & 73.3 & 59.4 & 8.9 & 10.5 & 29.1 & 52.1 & 27.0 & 39.5 & 39.1 & 2.9 & 74.8 & 76.5 & 58.1 & 11.7 & 2.6 & 15.8 & 4.5 & 14.9 & 6.5 & 10.9 & 48.3 & 15.6 & 26.5 \\
FGA-Seg \textcolor{black!62}{\tiny ArXiv'25} & 59.2 & \textbf{14.5} & 63.2 & 69.6 & 59.4 & 10.0 & \textbf{16.7} & 29.5 & 51.6 & 28.9 & 40.3 & \textbf{42.3} & 2.4 & 74.2 & 76.1 & 57.5 & \textbf{17.0} & 3.6 & \textbf{20.0} & 4.7 & \textbf{15.2} & 7.1 & 6.8 & 48.8 & 16.5 & 27.2 \\
OVRS \textcolor{black!62}{\tiny TGRS'25} & 59.5 & 10.3 & 65.6 & 73.4 & 60.0 & 12.3 & 13.7 & 30.0 & 52.2 & 29.0 & 40.6 & 40.2 & 2.4 & 74.6 & 77.8 & 58.5 & 14.6 & 1.9 & 13.4 & 7.8 & 13.5 & 7.3 & 18.7 & 48.8 & 17.0 & 27.6 \\
GSNet \textcolor{black!62}{\tiny AAAI'25} & 59.9 & 9.4 & 66.9 & 73.3 & 60.5 & 1.2 & 9.8 & 28.1 & 52.4 & 24.9 & 38.6 & 41.2 & 4.0 & 74.9 & 75.6 & \textbf{58.6} & 16.0 & 4.1 & 16.3 & 8.2 & 14.0 & 2.1 & 3.0 & 48.9 & 15.3 & 26.5 \\
RSKT-Seg \textcolor{black!62}{\tiny AAAI'26} & 58.8 & 10.3 & 66.5 & 74.8 & 60.0 & 9.0 & 12.3 & 29.8 & 52.6 & 27.8 & 40.2 & 42.1 & \textbf{4.8} & 77.2 & 76.8 & 58.3 & 16.1 & 4.2 & 14.3 & 9.5 & 13.9 & 4.3 & 17.4 & \textbf{50.2} & 17.3 & 28.2 \\
\rowcolor{yellow!25}
\textbf{GeoSeg-OV} & \textbf{60.8} & 9.1 & \textbf{70.9} & \textbf{75.2} & \textbf{61.6} & \textbf{24.0} & 15.9 & \textbf{34.4} & \textbf{54.0} & \textbf{34.0} & \textbf{44.0} & 35.0 & 2.1 & \textbf{79.1} & \textbf{78.5} & 57.9 & 14.5 & 6.8 & 13.8 & 11.8 & 14.9 & \textbf{9.5} & 27.2 & 48.7 & \textbf{19.6} & \textbf{29.2} \\
\bottomrule
\end{tabular}%
}
\end{table*}

\subsubsection{Auxiliary Vision Foundation Model Analysis}
\label{sec:vfm_ablation}

To verify that the proposed framework accommodates diverse auxiliary encoders, we evaluate three representative VFMs: DINOv2 ViT-B/14~\citep{oquab2023dinov2}, SAM 2.1 Hiera Base Plus~\citep{ravi2024sam}, and Depth Anything V2 ViT-B/14~\citep{yang2024depth}. All encoders remain entirely frozen and serve as inputs to SGA and CAD without any architecture modification.

\noindent\textbf{Framework generality.}
As shown in Table~\ref{tab:vfm-unified-heat}, all three VFMs yield strong performance, with average mIoU ranging from 43.8 to 44.2 (FLAIR) and from 40.8 to 41.6 (OpenEarthMap). The narrow gap across fundamentally different encoders confirms that the effectiveness of GeoSeg-OV stems from how auxiliary features are utilized rather than from a specific encoder choice.

\noindent\textbf{Per-VFM characteristics.}
Despite similar overall performance, each VFM exhibits strengths aligned with its pretraining objective. DINO excels on datasets with diverse land cover layouts (highest mIoU on LoveDA and EarthMiss under FLAIR training), suggesting that its self-supervised representations capture layout-level structural cues effectively. SAM performs strongly where fine-grained object boundaries are critical (highest mIoU on DeepGlobe under both settings; highest average mACC at 63.0/59.1), consistent with its boundary-sensitive pretraining. Depth Anything V2 achieves the highest average mIoU (44.2/41.6) and fwIoU (51.1/48.4) under both settings, with no notable weakness on any individual dataset.


\noindent\textbf{Default encoder selection.}
We select Depth Anything V2 as the default auxiliary encoder because it delivers the most balanced performance across datasets, whereas DINO and SAM exhibit greater performance variation across different scenarios. It also achieves the highest average mIoU (44.2/41.6) and fwIoU (51.1/48.4) under both training settings, demonstrating strong overall performance across both category-balanced and spatially dominant evaluations.

\begin{figure*}[t]
    \centering
    \includegraphics[width=0.95\textwidth]{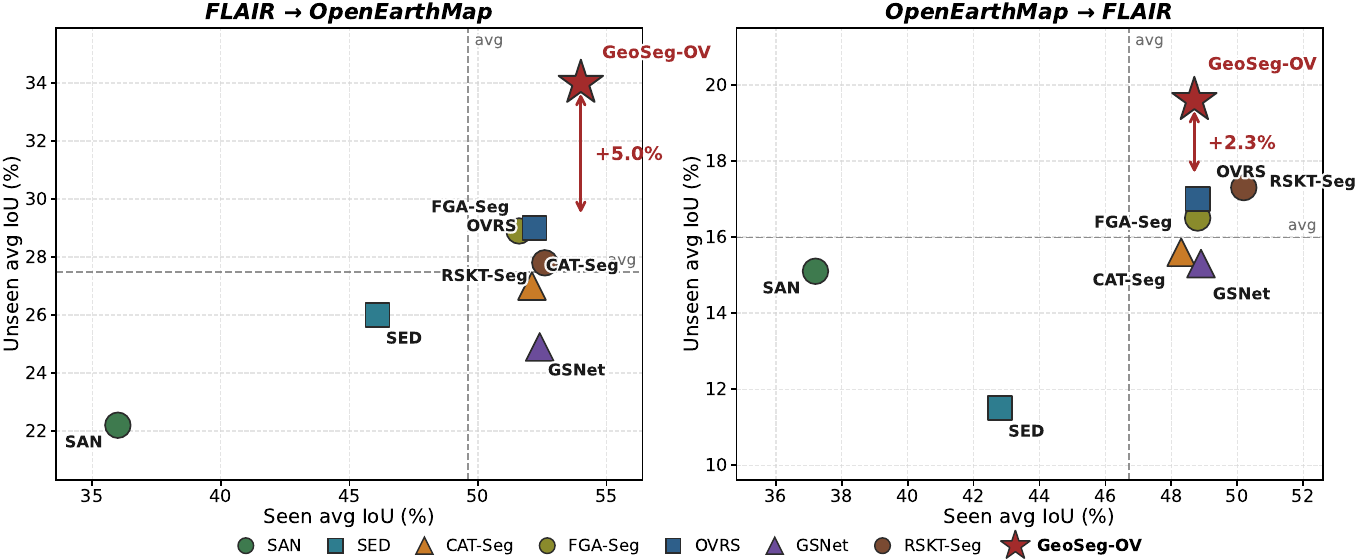}
    \caption{Seen vs.\ unseen class IoU for each method under cross-domain evaluation. Left: FLAIR $\rightarrow$ OpenEarthMap. Right: OpenEarthMap $\rightarrow$ FLAIR. Methods closer to the upper-right corner achieve better performance on both seen and unseen categories. GeoSeg-OV shows a strong seen--unseen trade-off in both directions.}
    \label{fig:seen_unseen_scatter}
\end{figure*}

\subsubsection{Open-Vocabulary Generalization Analysis}

To evaluate open-vocabulary generalization, we divide the evaluation categories into seen classes overlapping with the training vocabulary and unseen classes absent from it. The class-wise results are reported in Table~\ref{tab:classwise-flair-oem-combined}.

\noindent\textbf{FLAIR $\rightarrow$ OpenEarthMap.}
GeoSeg-OV achieves the highest overall, seen-class, and unseen-class IoUs of 44.0, 54.0, and 34.0, respectively. Its unseen-class performance exceeds the second-best OVRS by +5.0 IoU. In particular, GeoSeg-OV achieves 24.0 IoU on developed space, substantially outperforming GSNet, which achieves only 1.2 IoU. These results demonstrate that the category-agnostic structural guidance effectively complements visual--text matching when transferring to unseen categories.

\noindent\textbf{OpenEarthMap $\rightarrow$ FLAIR.}
This direction is more challenging because FLAIR contains eight unseen fine-grained categories with similar visual appearances, such as pervious versus impervious surfaces and coniferous versus deciduous trees. GeoSeg-OV achieves the best overall IoU of 29.2 and unseen-class IoU of 19.6, surpassing the second-best RSKT-Seg by +2.3 on unseen classes. Although its seen-class IoU of 48.7 is slightly lower than RSKT-Seg at 50.2, its clear advantage on unseen categories leads to the best overall performance.

\noindent\textbf{Seen--unseen trade-off.}
Fig.~\ref{fig:seen_unseen_scatter} jointly compares seen- and unseen-class performance, with stronger methods located closer to the upper-right corner. GeoSeg-OV achieves the most favorable overall balance in both transfer directions, demonstrating that structure-sensitive guidance improves generalization to unseen categories without compromising performance on the seen vocabulary.

\begin{figure}[!t]
    \centering
    \includegraphics[width=\linewidth]{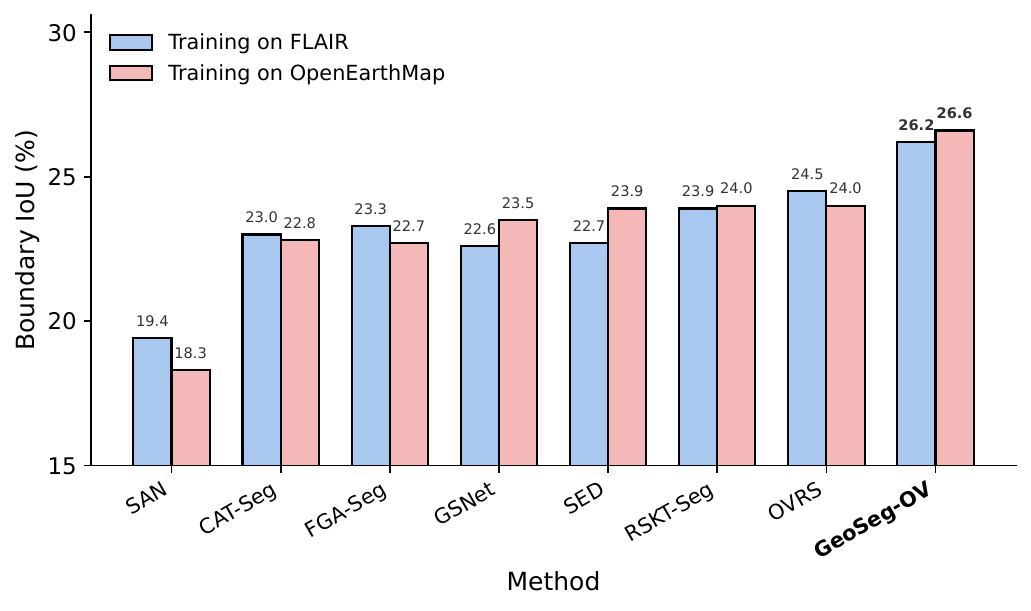}
    \caption{Boundary IoU comparison with different state-of-the-art methods, measuring the boundary-level agreement between predicted masks and ground-truth annotations.}
    \label{fig:boundary_iou}
\end{figure}




\subsubsection{Boundary Quality Analysis}

To assess boundary quality beyond region-level overlap, we report Boundary IoU in Fig.~\ref{fig:boundary_iou}, which measures the agreement between predicted and ground-truth boundary regions.

GeoSeg-OV achieves the highest Boundary IoU under both training settings, reaching 26.2\% when trained on FLAIR and 26.6\% when trained on OpenEarthMap, outperforming the corresponding second-best methods by +1.7\% and +2.6\%, respectively. These consistent gains demonstrate that structure-guided aggregation improves both region-level segmentation and boundary delineation, producing more accurate category transitions under cross-dataset domain shifts.

\begin{table}[!t]
\centering
\scriptsize
\definecolor{LTabHi}{RGB}{230, 245, 238}
\definecolor{LTabMid}{RGB}{240, 250, 245}
\caption{Efficiency and performance comparison of different methods. Runtime and memory are measured on a single NVIDIA RTX 4090 GPU.}
\label{tab:efficiency-performance}
\resizebox{\columnwidth}{!}{%
  \setlength{\tabcolsep}{1.5pt}%
  \renewcommand{\arraystretch}{1.06}%
  \begin{tabular}{@{}lccccc@{}}
    \toprule
    \textbf{Method} & Params.\,(M) & Training (s/it) & Inference (s/it) & Memory (GB) & mIoU \\
    \midrule
    SAN\,{\textcolor{black!62}{\fontsize{3.85}{4.2}\selectfont CVPR'23}} &
      157.8 & \textbf{0.17} & 0.05 & \textbf{3.3} & 26.9 \\
    SED\,{\textcolor{black!62}{\fontsize{3.85}{4.2}\selectfont CVPR'24}} &
      180.8 & 0.47 & \textbf{0.04} & 18.5 & 35.9 \\
    CAT-Seg\,{\textcolor{black!62}{\fontsize{3.85}{4.2}\selectfont CVPR'24}} &
      \textbf{154.5} & 0.26 & 0.13 & 8.5 & 39.3 \\
    FGA-Seg\,{\textcolor{black!62}{\fontsize{3.85}{4.2}\selectfont ArXiv'25}} &
      162.9 & 0.18 & 0.13 & 5.8 & 39.7 \\
    OVRS\,{\textcolor{black!62}{\fontsize{3.85}{4.2}\selectfont TGRS'25}} &
      \textbf{154.5} & 0.36 & 0.19 & 12.4 & 41.7 \\
    GSNet\,{\textcolor{black!62}{\fontsize{3.85}{4.2}\selectfont AAAI'25}} &
      244.2 & 0.37 & 0.27 & 11.4 & 38.9 \\
    RSKT-Seg\,{\textcolor{black!62}{\fontsize{3.85}{4.2}\selectfont AAAI'26}} &
      398.9 & 0.39 & 0.29 & 15.5 & 41.1 \\
    \rowcolor{LTabMid}%
    \textbf{GeoSeg-OV (w/o Rot)} &
      245.2 & 0.36 & 0.23 & 9.5 & 42.9 \\
    \rowcolor{LTabHi}%
    \textbf{GeoSeg-OV (Full)} &
      245.2 & 0.47 & 0.31 & 13.4 & \textbf{44.2} \\
    \bottomrule
  \end{tabular}%
}
\vspace{-0.4em}
\end{table}

\begin{figure*}[htbp]
    \centering
    \includegraphics[width=0.96\textwidth]{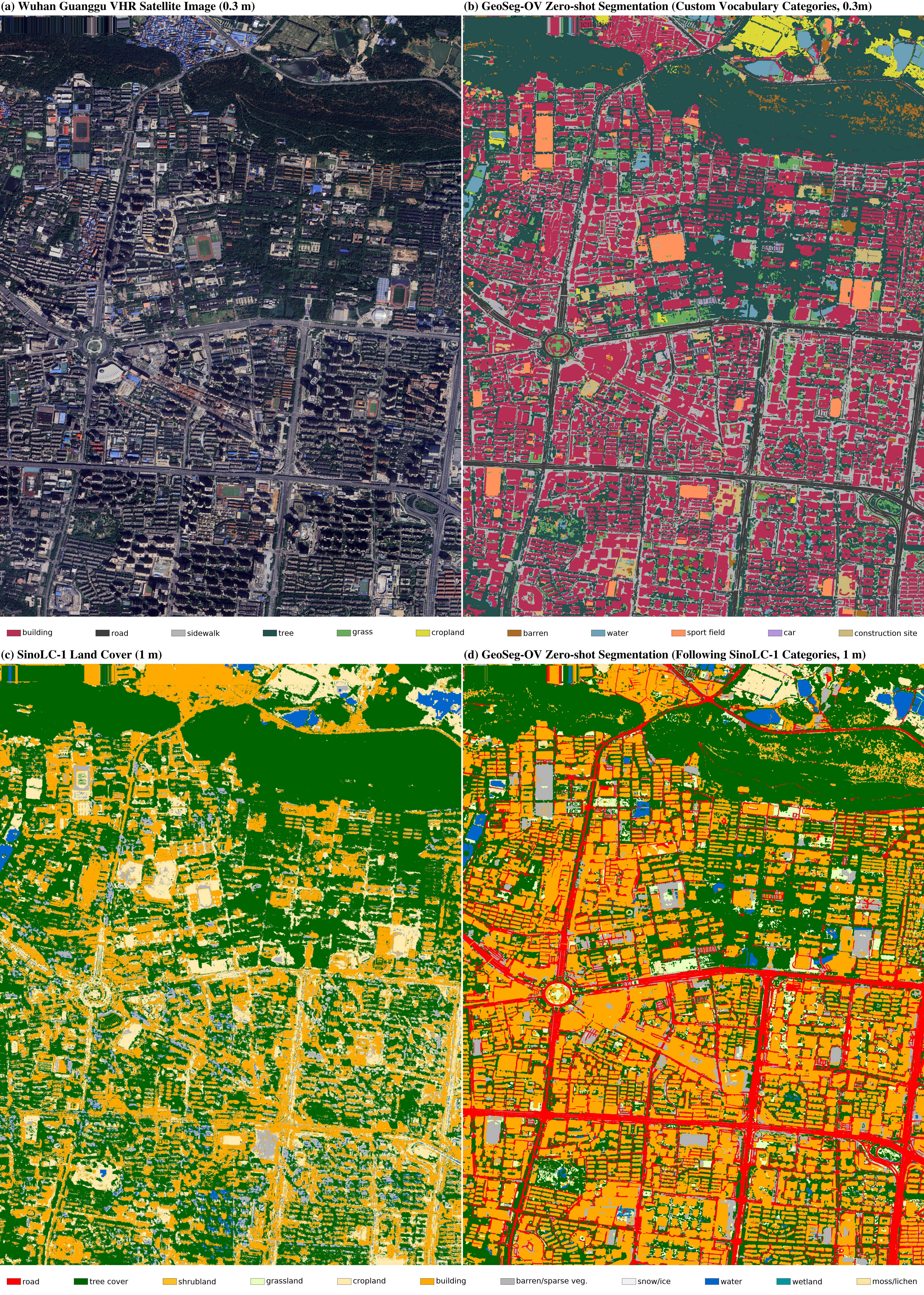}
    \caption{Large-scale zero-shot inference on Guanggu, Wuhan ($11{,}000 \times 15{,}000$+ pixels).
    \textbf{(a)}~VHR satellite image (0.3\,m).
    \textbf{(b)}~GeoSeg-OV with custom vocabulary.
    \textbf{(c)}~SinoLC-1 land-cover product (1\,m).
    \textbf{(d)}~GeoSeg-OV following SinoLC-1 categories.}
    \label{fig:large_scale}
\end{figure*}

\subsubsection{Efficiency Analysis}

Table~\ref{tab:efficiency-performance} compares efficiency and performance under the FLAIR training setting.

\noindent\textbf{Source of overhead.}
GeoSeg-OV introduces two sources of computational cost beyond CAT-Seg: a frozen auxiliary encoder forward pass, and the SGA attention bias computation together with the CAD refinement operations. The auxiliary encoder dominates the overhead; SGA adds only a lightweight projection and pairwise dot product within existing Swin windows, and CAD consists of depthwise separable convolutions. This is reflected in the comparison between GeoSeg-OV (w/o Rot) and CAT-Seg: the 0.10\,s/it increase in training time is primarily attributable to the auxiliary encoder, while SGA and CAD themselves add minimal latency. Multi-rotation encoding introduces a further 0.11\,s/it by requiring four CLIP forward passes instead of one.

\noindent\textbf{Accuracy-efficiency trade-off.}
GeoSeg-OV (Full) achieves 44.2 mIoU at 0.47\,s/it training and 0.31\,s/it inference. The variant without rotation achieves 42.9 mIoU, already surpassing all baselines, at 0.36\,s/it training, 0.23\,s/it inference, and 9.5\,GB memory. Thus, adding the frozen encoder with SGA and CAD yields +3.6 mIoU over CAT-Seg at only +0.10\,s/it training cost---a substantially better accuracy-per-compute ratio than multi-rotation encoding (+1.3 mIoU at +0.11\,s/it) or the auxiliary matching streams employed by GSNet and RSKT-Seg.

\noindent\textbf{Comparison with AVTM methods.}
RSKT-Seg requires two auxiliary encoders (RemoteCLIP + DINO) that both participate in cost-map construction, resulting in 398.9M parameters and 0.29\,s/it inference for 41.1 mIoU. GeoSeg-OV uses a single frozen encoder whose features only produce a lightweight attention bias and guidance refinement signal, yielding comparable inference time (0.31\,s/it) at 62\% of the parameters while achieving +3.1 higher mIoU. GSNet similarly employs an auxiliary encoder for cost-map construction (244.2M, 0.27\,s/it) but reaches only 38.9 mIoU, further indicating that using structural priors as guidance achieves a better accuracy-efficiency balance than auxiliary matching streams.

\subsection{Case Study on Large-Scale Zero-Shot Transfer}

To evaluate the practical applicability of GeoSeg-OV beyond benchmark datasets, we apply the model trained on OpenEarthMap directly to a very-high-resolution satellite image covering the Guanggu district of Wuhan, China. The image contains more than $11{,}000\times15{,}000$ pixels with a ground sampling distance of 0.3\,m as shown in Fig.~\ref{fig:large_scale} (a). GeoSeg-OV performs sliding-window inference over the full-resolution image without using any annotations from the target scene or conducting target-domain fine-tuning.

We first evaluate the flexibility of GeoSeg-OV using a custom vocabulary for fine-grained urban mapping, including building footprints, road surfaces, tree canopy, water bodies, bare soil, and impervious surfaces. As shown in Fig.~\ref{fig:large_scale} (b), GeoSeg-OV produces a spatially coherent land-cover map over the entire scene. In particular, road networks remain continuous across large spatial extents, building regions exhibit well-defined footprints, and vegetation areas preserve coherent spatial distributions. These results demonstrate the ability of GeoSeg-OV to transfer from the OpenEarthMap training domain to a large-scale urban scene while supporting task-specific category specification without additional annotations or model adaptation. This flexibility provides considerable potential for applications such as urban morphology analysis, impervious-surface mapping, green-space assessment, and transportation infrastructure extraction.

We further evaluate cross-vocabulary transfer by replacing the custom vocabulary with the category schema of SinoLC-1~\citep{li2023sinolc}, a widely used 1\,m land-cover product derived from supervised classification. For visual comparison, both the GeoSeg-OV prediction and SinoLC-1 are presented at a spatial resolution of 1\,m in Fig.~\ref{fig:large_scale} (c--d). Despite having never been trained on SinoLC-1 data, GeoSeg-OV directly produces predictions under the new category schema through text-based category specification. Compared with SinoLC-1, the GeoSeg-OV result exhibits visually clearer building regions, more continuous road structures, and finer spatial transitions between adjacent land-cover categories. Although this comparison is qualitative because pixel-level reference annotations are unavailable for the target scene, it highlights the potential of GeoSeg-OV to generate detailed land-cover maps under the SinoLC-1 category schema without retraining. Overall, these results demonstrate the ability of GeoSeg-OV to generalize across geographic domains and category systems without target-domain annotations or model retraining, highlighting its potential as a flexible framework for large-scale open-vocabulary land-cover mapping.

\section{Conclusion}

We presented GeoSeg-OV, a structure-guided framework that integrates structural priors into cost aggregation and progressive decoding. GeoSeg-OV introduces a new paradigm for utilizing auxiliary VFMs by decoupling their features from visual--text matching and repurposing them as structural guidance. SGA jointly integrates cost tokens and CLIP semantic guidance with VFM-derived pairwise structural biases to produce spatially coherent and semantically discriminative cost representations, followed by text-conditioned class-wise reasoning to model inter-category relationships. CAD further adapts multi-scale semantic and structural guidance according to the current decoder context for progressive prediction. We established a global HRLC benchmark comprising seven datasets across six continents for rigorous cross-dataset evaluation. Extensive experiments demonstrate that GeoSeg-OV achieves state-of-the-art average performance under substantial resolution and geographic shifts. The large-scale case study further highlights its practical potential for flexible open-vocabulary land-cover mapping in real-world remote sensing applications.

\printcredits

\bibliographystyle{cas-model2-names}

\bibliography{cas-refs}


\end{document}